\documentclass[runningheads]{llncs}

\usepackage{eccv}

\usepackage{eccvabbrv}

\usepackage{graphicx}
\usepackage{booktabs}
\usepackage{amsmath}
\usepackage{bbm}
\usepackage{pifont}
\usepackage{adjustbox}
\usepackage{multirow}
\usepackage{algorithm}
\usepackage{algpseudocode}
\newcommand{\cmark}{\ding{51}}%
\newcommand{\xmark}{\ding{55}}%

\DeclareMathOperator*{\argmax}{arg\,max}
\DeclareMathOperator*{\argmin}{arg\,min}
\newcommand{\newheader}[1]{{\noindent \textbf{#1} \hspace{0.25em}}}
\usepackage[accsupp]{axessibility}  

\usepackage{hyperref}

\usepackage{orcidlink}

\newcommand{\camera}[1]{#1}

\begin{document}

\title{Efficient Bias Mitigation Without Privileged Information}

\titlerunning{Efficient Bias Mitigation Without Privileged Information} 

\author{Mateo Espinosa Zarlenga\thanks{Work done while the author was an intern at Sony AI.}\inst{1}\orcidlink{0009-0006-7333-5727}\and
Swami Sankaranarayanan\inst{2}\orcidlink{0000-0002-9239-3221} \and
Jerone T. A. Andrews\inst{2}\orcidlink{0000-0002-8552-1213} \and
Zohreh Shams\inst{1}\orcidlink{0000-0002-0143-798X} \and
Mateja Jamnik\inst{1}\orcidlink{0000-0003-2772-2532} \and 
Alice Xiang\inst{2}\orcidlink{0000-0001-7907-9353}
}

\authorrunning{M.~Espinosa~Zarlenga et al.}

\institute{
University of Cambridge \\
\email{\{me466, zs315\}@cam.ac.uk, mateja.jamnik@cl.cam.ac.uk}
\and
Sony AI \\
\email{\{swami.sankaranarayanan, jerone.andrews, alice.xiang\}@sony.com}\\
}

\maketitle

\begin{abstract}
    Deep neural networks trained via empirical risk minimization often exhibit significant performance disparities across groups, particularly when group and task labels are spuriously correlated (e.g., ``grassy background'' and ``cows'').
    Existing bias mitigation methods that aim to address this issue often either rely on group labels for training or validation, or require an extensive hyperparameter search.
    Such data and computational requirements hinder the practical deployment of these methods, especially when datasets are too large to be group-annotated, computational resources are limited, and models are trained through already complex pipelines.
    In this paper, we propose Targeted Augmentations for Bias Mitigation (TAB), a simple hyperparameter-free framework that leverages the \textit{entire} training history of a helper model to identify spurious samples, and \emph{generate} a group-balanced training set from which a robust model can be trained.
    We show that TAB improves worst-group performance without \textit{any} group information or model selection, outperforming existing methods while maintaining overall accuracy.
\end{abstract}


\section{Introduction}
\label{sec:introduction}

It is often difficult to place the resounding success of Deep Neural Networks (DNNs)~\cite{silver_alphago_nature_16, vaswani_attention_neurips_17, ho_diffision_models_neurips_20, radford_clip_icml_21} in conjunction with their tendency to exploit biases within their training sets~\cite{buolamwini_gender_shades_facct_18, hendricks_overcoming_bias_in_captioning_models_spurious_correlations_eccv_18, geirhos_shortcut_learning_spurious_correlations_nature_20, oakden_hidden_stratification_medical_imaging_bias_20, degrave_bad_ml_covid_shortcut_nature_21}.
Nevertheless, in visual datasets, arguably the driving force of several consumer-facing AI models, such biases have been long known to be exploited and amplified by models trained on their data~\cite{zhao_reducing_bias_amplification_spurious_correlations_emnlp_17, wang_bias_amplification_balanced_datasets_spurious_correlations_iccv_19}.
The difficulty of addressing these biases lies in their ubiquity and diversity: biases may arise from spurious correlations between salient attributes (e.g., ``water background'') and classification task labels (e.g., ``boat'') ~\cite{torralba_unbiased_look_at_dataset_bias_cvpr_11, singh_learning_contextual_bias_spurious_correlation_cvpr_20, geirhos_shortcut_learning_spurious_correlations_nature_20, degrave_bad_ml_covid_shortcut_nature_21}, as well as from underrepresentation of diverse geographical/cultural artifacts~\cite{devries_object_recognition_for_everyone_dataset_bias_cvpr_19, wang_revise_visual_bias_discovery_22}, gender-related attributes~\cite{kay_unequal_gender_representaion_dataset_bias_15, wang_directional_bias_amplification_spurious_correlations_icml_21, zhao_multi_attribute_bias_amplification_spurious_correlations_icml_23}, object subcategories~\cite{zhu_long_tail_object_subcategory_dataset_bias_cvpr_14}, and skin tones~\cite{buolamwini_gender_shades_facct_18}.
Such multi-source provenance of dataset biases begs one to question the ethical~\cite{oneal_weapons_of_math_destruction_black_box_ethics, rudin2021interpretable} and legal~\cite{yu_law_black_box_ethics_19, goodman_accountable_gdpr_black_box_law_ethics_neurips_16} ramifications of blindly deploying DNNs in critical tasks where biases may be unknowingly perpetuated by the model.

As the AI community has become aware of the effect of such biases on data-driven models (e.g., DNNs), the field has experienced a surge of bias mitigation~\cite{sagawa_gdro_bias_mitigation_19, levycvar_dro_bias_mitigation_neurips_20, sohoni_george_bias_mitigation_neurips_20, nam_lff_bias_mitigation_neurips_20, liu_jtt_bias_mitigation_icml_21} and bias discovery~\cite{wu_errudite_visual_bias_discovery_19, wang_revise_visual_bias_discovery_22, deon_spotlight_bias_discovery_facct_22, kim_bias_to_text_bias_discovery_23, yenamandra_facts_bias_discovery_iccv_23, eyuboglu_domino_bias_discovery_iclr_23} pipelines.
By studying bias in the form of spurious correlations, these methods aim to circumvent the difficulty of vetting large datasets for task-irrelevant correlations~\cite{andrews_ethical_considerations_dataset_bias_icml_23} and quality-control~\cite{prabhu_large_image_dataset_bias_20, birhane_multimodal_dataset_bias_21, crawford_excavating_ai_politics_image_dataset_bias_21}.
However, most previous methods require \textit{privileged information} in the form of group labels (e.g., ``water background'', ``land background'', etc.) either during training, as additional information for the loss function, or during validation, for model selection.
These constraints, together with the hyperparameter sensitivity~\cite{liu_jtt_bias_mitigation_icml_21, tsirigotis_no_labels_ssl_bias_mitigation_neurips_23} and the fairness-accuracy trade-off~\cite{ma_fairness_accuracy_tradeoff_neurips_22, li2023_whac_a_mole_shorcuts_amplification} in existing methods, have rendered most existing mitigation pipelines difficult, if not impossible, to apply to real-world setups where biases are unknown and hyperparameter searches are prohibitively expensive.

\begin{figure}[t!]
    \centering\includegraphics[width=\textwidth]{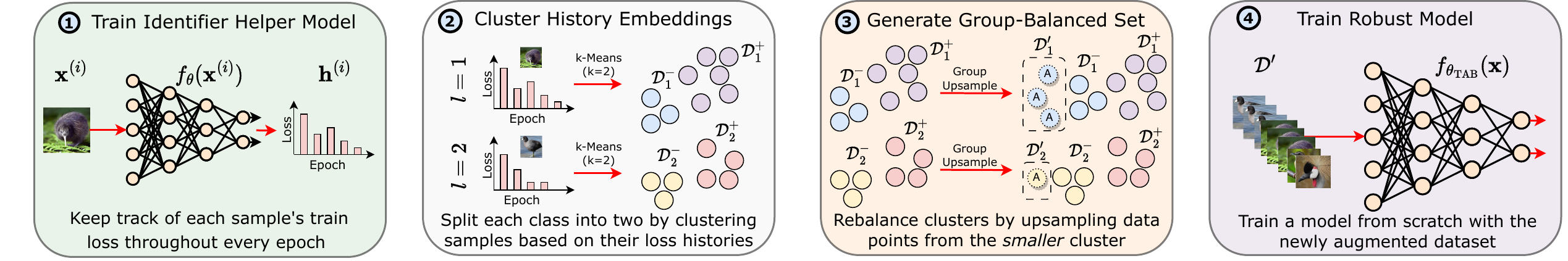}
    \caption{
        Given a model $f_\theta$ and a training set $\mathcal{D}$, TAB learns a robust model as follows:
        (1) We train $f_\theta$ while keeping track of each sample's loss at every epoch.
        (2) We split samples of each class $l$ into two groups, a minority group $\mathcal{D}^{-}_l$ and a majority group $\mathcal{D}^{+}_l$, by clustering samples using their \textit{entire} training histories.
        (3) We construct a group-balanced dataset by upsampling each minority group $\mathcal{D}_l^-$ so that $\mathcal{D}_l^-$ has the same size as the majority group $\mathcal{D}_l^+$. 
        (4) We train $f_\theta$ from scratch using the augmented dataset.
    }
    \label{fig:abstract}
\end{figure}

In this paper, we address this gap by studying group-unsupervised bias mitigation in vision models from a practical perspective.
We focus on vision models given their relevance to modern architectures~\cite{bommasani_foundation_models_21, ramesh_zero_shot_image_gen_icml_21, he2022masked} and critical applications~\cite{badue2021self}, as well as their proclivity to propagate biases~\cite{tommasi_deeper_look_at_dataset_bias_17, fabbrizzi2022survey}.
Specifically, we propose \textit{Targeted Augmentations for Bias mitigation (TAB)}, a simple yet effective hyperparameter-free unsupervised bias mitigation pipeline (Figure~\ref{fig:abstract}).
Inspired by recent work showing the effectiveness of group-balanced training for improving robustness~\cite{kirichenko_last_layer_retraining_bias_mitigation_iclr_23, labonte_last_layer_retraining_neurips_23}, our approach exploits the \textit{entire} training history of a helper model to \textit{generate} a group-balanced dataset from which a robust model can be trained.
The novel use of a model's entire training dynamics for rebalancing enables TAB to significantly improve worst-group accuracy without (a) any group information, (b) an expensive hyperparameter search, (c) significant sacrifices to mean accuracy, or (d) considerable changes to existing training pipelines, making TAB applicable to real-world tasks.

Our work's main contributions can, therefore, be summarised as follows:
\begin{enumerate}
    \item We highlight that in the absence of \textit{any} group information, many bias mitigation methods may fail to mitigate known biases due to the inherent difficulty of model selection, incurring a \textit{price for their bias unawareness}.   
    
    \item We introduce \textit{Targeted Augmentations for Bias mitigation} (TAB), an efficient hyperparameter-free bias mitigation pipeline that does not require group annotations for training or validation.\footnote{Our code is available at \href{https://github.com/SonyResearch/tab_bias_mitigation}{https://github.com/SonyResearch/tab\_bias\_mitigation}.}

    \item We show that TAB achieves better worst-group performance than previous unsupervised methods without significant sacrifices to mean accuracy or
    an expensive hyperparameter search.
    
\end{enumerate}


\section{Problem Formulation}
\label{sec:background_formulation}

\newheader{Setup} In this paper, we study bias in the form of spurious correlations within supervised learning.
We assume we have a training set $\mathcal{D}_\text{train} := \{(\mathbf{x}^{(i)}, y^{(i)}) \}_{i=1}^N$ where each sample $\mathbf{x}^{(i)} \in \mathbb{R}^n$ and label $y^{(i)} \in \{1, \cdots, L\}$ are sampled from a distribution $\mathcal{P}_\text{train}(\mathbf{x}, y)$.
Our goal is to learn the parameters $\theta \in \Theta$ of a DNN $f_\theta: \mathbb{R}^n \rightarrow \{1, \cdots, L\}$ so that $f_\theta$ maps each sample $\mathbf{x}^{(i)}$ to its label $y^{(i)}$.

Traditionally, $\theta$ is learned by minimising the expected value of a loss function $\ell(f_\theta(\mathbf{x}^{(i)}), y^{(i)})$, which reflects how accurately $f_\theta(\mathbf{x}^{(i)})$ approximates $y^{(i)}$.
As $\mathcal{P}_\text{train}$ is generally unknown in practice, given a finite training set $\mathcal{D}_\text{train}$, the optimal parameters $\theta^*$ are usually estimated by optimising the following \textit{Empirical Risk Minimisation} (ERM)~\cite{vapnik_erm_91} objective via gradient descent:
\begin{align}
    \theta^* = \argmin_{\theta \in \Theta} \mathbb{E}_{(\mathbf{x}, y) \sim \mathcal{P}_\text{train}(\mathbf{x}, y)} \big[ \ell(f_\theta(\mathbf{x}), y) \big] \approx \argmin_{\theta \in \Theta} \sum_{i=1}^N \frac{\ell\big(f_\theta(\mathbf{x}^{(i)}), y^{(i)}\big)}{N}
\end{align}

\indent\newheader{Worst-group Performance}
When spurious correlations exist between irrelevant yet salient features (e.g., ``grass’’) and downstream labels (e.g., ``cow’’), ERM is known to encourage the exploitation of such ``shortcuts’’, resulting in the unfair treatment of \textit{bias-conflicting} samples---those lacking the spurious correlation~\cite{nam_lff_bias_mitigation_neurips_20, geirhos_shortcut_learning_spurious_correlations_nature_20, degrave_bad_ml_covid_shortcut_nature_21}.
Such behaviour can go unnoticed when using traditional population-level evaluation metrics (e.g., average accuracy).
For example, in an object detection dataset where $98\%$ of images depict ``cows'' on ``grass'', a model could achieve a mean accuracy of $98\%$ by predicting ``cows'' whenever ``grass'', a salient feature, appears.
This statistic, however, hides the model's inability to detect cows on diverse backgrounds, such as ``roads'', where its accuracy could be $0\%$, rendering it unfit for deployment in, say, self-driving car pipelines.

Therefore, we study the performance of DNNs across all population subgroups.
That is, we assume every sample $(\mathbf{x}^{(i)}, y^{(i)})$ belongs to one of $k$ groups $g^{(i)} \in \{1, \cdots, k\}$, and analyse models through their \textit{worst-group accuracy} (WGA):
\begin{align}
    \text{WGA}\big(f_\theta, \mathcal{P} \big) := \min_{g \in \{1, 2, \cdots, k\}} \mathbb{E}_{(\mathbf{x}, y) \sim \mathcal{P}(\mathbf{x}, y | g)} \big[ \mathbbm{1}\big(f_\theta(\mathbf{x}) = y \big)\big]
\end{align}
where $\mathbbm{1}(\cdot)$ is the indicator function and $\mathcal{P}(\mathbf{x}, y | g)$ is the joint distribution of samples in group $g \in \{1, \cdots, k\}$.
We note that we assume knowledge of group identities only during test time and \emph{not} for training or validation.
With these labels, we can compute test WGA via the empirical worst-group mean accuracy.


\section{Related Work}
\label{sec:background}
Previous research studying the effect of spurious correlations on DNNs falls into two camps: (1) \textit{bias mitigation} (\textit{BM}), and (2) \textit{bias discovery} (BD).
Here, we first discuss BM methods, separating them between supervised and unsupervised approaches, and then discuss related BD works.

\indent\newheader{Group-supervised BM}
When group annotations are available (i.e., we know each sample's subgroup), group-supervised BM methods improve a DNN's worst-group performance by leveraging these labels during training.
Distributionally robust optimisation (DRO) methods~\cite{ben_dro, duchi_dro, sagawa_gdro_bias_mitigation_19, wei_posthoc_dro_iclr_23}, for example, introduce a group-aware loss function to learn models that are robust to distribution shifts.
Within these approaches, Group DRO (G-DRO)~\cite{hu_gdro_og, oren_gdro_language, sagawa_gdro_bias_mitigation_19} is typically used to train oracle-like unbiased models, minimising the empirical worst-group training loss.
Other works instead propose reweighting~\cite{ren_learning_to_reweight_icml_18} or rebalancing~\cite{kirichenko_last_layer_retraining_bias_mitigation_iclr_23, labonte_last_layer_retraining_neurips_23, wang_overwriting_pretrained_bias_iccv_23} samples in the training distribution, assigning larger weights to underrepresented subgroups. 
In contrast, representation learning approaches, such as adversarial methods~\cite{zhang_adversarial_bias_mitigation_18, ramakrishnan_adversarial_vqa_bias_mitigation_neurips_18} and invariant representation learning~\cite{ahmed_pgi_bias_mitigation_icml_20}, propose learning robust representations from which protected attributes (e.g., ``gender'') cannot be predicted.
Finally, recent re-annotation methods explore using a small group-annotated set to generate pseudo-labels for a larger annotated set~\cite{nam2022_spread_bias_mitigation_iclr_22}.

\indent\newheader{Group-unsupervised BM} The need for group annotations during training renders group-supervised BM approaches impractical for real-world workloads where biases are unknown and annotations are prohibitively expensive.
Several group-unsupervised BM methods have been proposed to circumvent this limiting requirement.
Promising directions include: (1)~discovering pseudo-group labels to run a group-supervised pipeline~\cite{sohoni_george_bias_mitigation_neurips_20, bahng_debias_reprs_from_biased_icml_20,  nam_lff_bias_mitigation_neurips_20, liu_jtt_bias_mitigation_icml_21, creager_eiil_icml_21}; (2)~introducing a loss that incentivises group-robustness~\cite{levycvar_dro_bias_mitigation_neurips_20, pezeshki_gradient_starvation_bias_mitigation_neurips_21}; and (3)~learning a rebalancing of the data to perform group-balanced retraining~\cite{liu_jtt_bias_mitigation_icml_21, labonte_last_layer_retraining_neurips_23}.
Other approaches involve constructing ensembles for learning robust classifiers~\cite{kim_multiaccuracy_bias_mitigation_19}, or dynamically labelling pseudo-groups through adversarial training~\cite{li_debian_bias_mitigation_eccv_22, paranjape_agro_bias_mitigation_iclr_23}.
More recent works have explored self-supervised pretraining pipelines for robust representation learning~\cite{kulkarni_multitask_ssl_bias_mitigation_tmlr_24}, and introduced neural architectures with robustness-aware inductive biases~\cite{taghanaki_cim_bias_mitigation_icml_21, shrestha_occamnets_bias_mitigation_eccv_22, asgari_masktune_bias_mitigation_neurips_22}.

Among these works, the methods closest to ours are ``Just Train Twice'' (JTT)~\cite{liu_jtt_bias_mitigation_icml_21} and ``Learning from Failure'' (LfF)~\cite{nam_lff_bias_mitigation_neurips_20}, as they both use an ``identifier'' model's output to identify bias-conflicting samples.
JTT achieves this through a two-pass solution where an identifier model $f^{(\text{id})}_{\theta_\text{id}}$ is trained for a small number of epochs $T$, and then $f_\theta$ is trained from scratch on a training set where samples mispredicted by $f^{(\text{id})}_{\theta_\text{id}}$ are upweighted by a factor of $\lambda_\text{up}$ (a hyperparameter).
In contrast, LfF jointly learns a bias identifier $f^{(\text{id})}_{\theta_\text{id}}$ and a debiased model $f_\theta$ by dynamically weighting $f_\theta$'s samples proportionally to the biased model's loss.

Concurrent with our work, two unsupervised BM methods, SELF~\cite{labonte_last_layer_retraining_neurips_23} and uLA~\cite{tsirigotis_no_labels_ssl_bias_mitigation_neurips_23}, also use an identifier model.
Our work distinguishes itself from these and all prior works by taking a practitioner-first approach and avoiding the need for (i) hyperparameters, (ii) complex pipelines, and (iii) train or validation group annotations (for a summary of related work, refer to Table~\ref{table:previous_bias_mitigation} in Appendix~\ref{appendix:previous_work}).
This difference, as discussed next, is crucial for advancing towards practical BM.

\indent\newheader{Bias Discovery}
\label{sec:background_bias_discovery}
A research direction related to unsupervised BM is bias discovery (BD), where one analyses biases in datasets via a model's behaviour.
Recent prominent methods, including 
FACTS~\cite{yenamandra_facts_bias_discovery_iccv_23}, Spotlight~\cite{deon_spotlight_bias_discovery_facct_22}, and Domino~\cite{eyuboglu_domino_bias_discovery_iclr_23}, partition the training set into distinct subgroups via error-aware models.
Other lines of work, such as Errudite~\cite{wu_errudite_visual_bias_discovery_19} and REVISE~\cite{wang_revise_visual_bias_discovery_22}, instead discover biases through human-in-the-loop visualisations that highlight error data slices.
More recently, Bias-to-Text~\cite{kim_bias_to_text_bias_discovery_23} has explored leveraging off-the-shelf vision-language models~\cite{radford_clip_icml_21} to describe biased slices using natural language.
Although all BD methods are useful for BM after biases have been discovered, their primary focus lies in discovering underlying biases within datasets rather than mitigating biases during training.
As a result, we focus entirely on framing our proposed method in the context of unsupervised BM methods.


\section{The Price of Unawareness}
\label{sec:importance_val_data}

In this paper, we are interested in exploring bias mitigation without privileged information such as group labels during training or validation.
Here, we discuss why this task is particularly challenging, highlighting the inherent difficulty of model selection when validation group labels are absent.
For this, we analyse the training dynamics of a ResNet-34~\cite{resnet} model trained on \texttt{Waterbirds}~\cite{hu_gdro_og}, a bird detection task with known spurious correlations (see \S\ref{sec:results}),
to illustrate why mean accuracy-based model selection is prone to selecting suboptimal hyperparameters for BM methods (Figure~\ref{fig:training_loss_curves}).
We use our observations to first frame this difficulty as a \textit{cost paid by BM pipelines requiring model selection} when they are unaware of biases in their validation sets and then motivate our method in the next section.

\begin{figure}[t!]
    \centering\includegraphics[width=\textwidth]{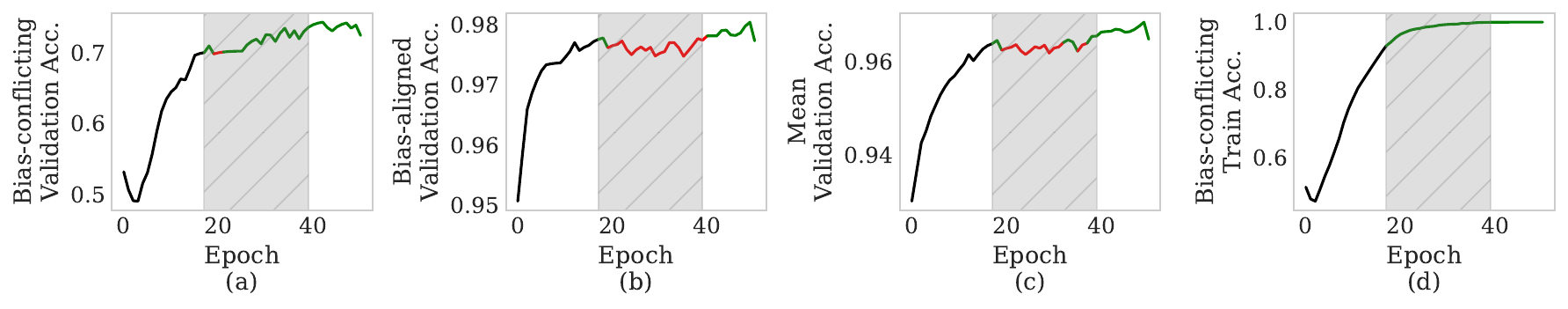}
    
    \caption{
        Accuracy throughout training on \texttt{Waterbirds} for (a) bias-conflicting validation samples, (b) bias-aligned validation samples, (c) all validation samples, and (d) bias-conflicting training samples.
        The shaded area marks when, as the loss converges (plot d), the validation WGA increases (green segment in plot~a) while the bias-aligned accuracy (plot~b) and mean accuracy (plot~c) temporarily decrease (red segments).
    }
    \label{fig:validation_dynamics}
\end{figure}

\indent\newheader{Model selection without groups}
Previous work within BM has referenced the difficulty of model selection without group annotations in the held-out validation set~\cite{liu_jtt_bias_mitigation_icml_21, asgari_masktune_bias_mitigation_neurips_22, tsirigotis_no_labels_ssl_bias_mitigation_neurips_23}.
We argue that this is due to a combination of two factors: 

\begin{enumerate}
    \item \textbf{Overfitting to validation biases.} When the validation set's distribution, $\mathcal{P}_\text{val}$, mirrors that of the training set, $\mathcal{P}_\text{train}$, as commonly seen in training pipelines, any spurious correlations in $\mathcal{P}_\text{train}$ persist in $\mathcal{P}_\text{val}$. Thus, models exploiting these correlations to achieve low empirical risk during training will similarly exhibit low error rates on the equally biased validation set.

    \item \textbf{Validation biases can mislead early stopping.} Since DNNs have a bias towards learning simpler hypotheses~\cite{valle_dnn_biased_towards_simpler_functions_iclr_19, shah_simplicity_bias_neurips_20},
    they are prone to learning spurious samples before harder bias-conflicting samples~\cite{arpit_dnn_memorisation_icml_17, nam_lff_bias_mitigation_neurips_20}
    (a fact exploited by several unsupervised BM methods~\cite{nam_lff_bias_mitigation_neurips_20, liu_jtt_bias_mitigation_icml_21, creager_eiil_icml_21, labonte_last_layer_retraining_neurips_23}).
    This tendency implies that for the model to truly generalise, it must first \textit{unlearn to exploit spurious correlations}. As seen in Figure~\ref{fig:validation_dynamics}, such unlearning results in a period during training where the number of correctly predicted bias-conflicting validation samples continues to increase (shaded area in Figure~\ref{fig:validation_dynamics}\textcolor{red}{a}), while the bias-aligned accuracy decreases for a significant number of epochs (Figure~\ref{fig:validation_dynamics}\textcolor{red}{b}), before increasing again.
    This decrease in bias-aligned accuracy, a majority group in the dataset, can lead to a \textit{temporary} decrease in average accuracy (Figure~\ref{fig:validation_dynamics}\textcolor{red}{c}), therefore yielding a model with worse average validation accuracy if training stops in this interpolation regime (as it would when using \textit{early-stopping}, a common component of modern pipelines).
\end{enumerate}

\begin{figure}[t!]
    \centering\includegraphics[width=\textwidth]{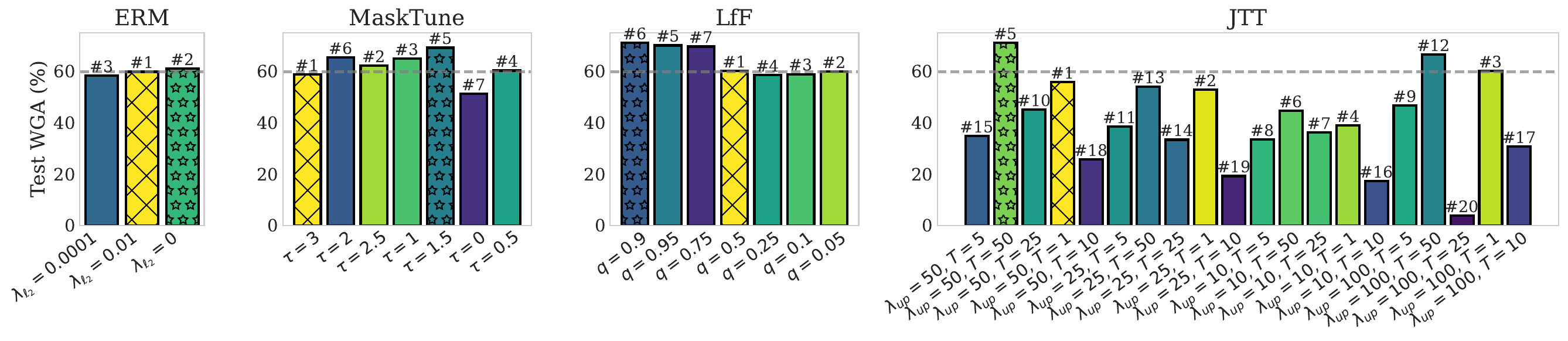}
     
     \caption{
        Test WGA for several hyperparameterisations of ERM, MaskTune~\cite{asgari_masktune_bias_mitigation_neurips_22},  LfF~\cite{nam_lff_bias_mitigation_neurips_20}, and JTT~\cite{liu_jtt_bias_mitigation_icml_21} on \texttt{Waterbirds}.
        The crossed yellow bar shows the model with the highest validation accuracy across each method.
        This model has (1) a lower WGA than the model selected using validation WGA (starred bar), and (2) a similar or worse WGA than an equivalent ERM model (dashed line), implying the selected models are as biased as their ERM equivalents (i.e., no debiasing occurred).
        For clarity, on top of each bar we show each model's ranking when ordered based on their validation accuracies.
    }
    \label{fig:model_selection}
\end{figure}

\noindent Together, these factors imply that mean accuracy-based model selection is prone to rejecting models that generalise better since these models fail to achieve a high validation accuracy when they do not fully exploit underlying biases.
Hence, if there exists a set of hyperparameters that enable a BM method to learn an ERM-like hypothesis, such a hypothesis will likely result in a higher validation accuracy than competing hyperparameters and, thus, will be selected during model selection.
As shown in Figure~\ref{fig:model_selection}, we see these factors manifesting themselves in real-world scenarios where poorly generalised models are chosen over better counterparts when hyperparameters are selected based on mean accuracy.
This difficulty has led to several unsupervised BM methods requiring group-balanced or group-annotated validation sets for model selection
(see Table~B.1 in App.~B).

\indent\newheader{The cost of validation bias unawareness}
The importance of the validation set in existing BM pipelines is a crucial yet often ignored aspect of these algorithms.
Nevertheless, given that unsupervised BM approaches aim to be utilised in settings where group identities are unknown, understanding the performance of a specific method in the absence of group labels is paramount.
Here, we study this via two scores.
The first score, which we call the \textit{Price of Unawareness} (PoU), measures the cost of lacking group labels for model selection.
This is captured as the ratio between
the best theoretically attainable test WGA for the BM pipeline vs the test WGA attained when hyperparameters are selected based on the validation mean accuracy.
When group labels are available during \textit{evaluation}, we can estimate this score by iterating over a set of representative candidate hyperparameters $\Gamma$ and computing the ratio between the best test WGA over all candidates and the test WGA of the candidate with the highest validation accuracy.
Intuitively, this score indicates how much one ``pays'' in the worst case when performing model selection for a BM pipeline without any group information.
Hence, a PoU value greater than $1$ indicates that the absence of validation group annotations leads to a model with worse WGA than what the BM method could have achieved had it used a proper group-annotated validation set.
In contrast, an \textit{ideal PoU} of $1$ indicates that a method's WGA is independent of whether WGA or accuracy is used for model selection.

The second metric we use looks at the \textit{mean model selection} (MMS) WGA of a given BM method by averaging the WGAs obtained for multiple hyperparameter candidates $\gamma \in \Gamma$.
This metric captures a BM method's \textit{expected behaviour} if one does not have the compute and time budgets required to run several hyperparameters.
We formally define both of these scores in Appendix~\ref{appendix:pou}.


\section{Targeted Augmentations}
\label{sec:tab}

In the previous section, we observed that existing BM approaches may fail to address biases in the absence of group information during model selection.
Specifically, we argued that this is an inherent limitation of pipelines that involve several hyperparameters or where the number of hyperparameters leading to an ERM-like hypothesis is high. This limitation underscores the importance of considering the type and number of hyperparameters when designing BM pipelines.
In this section, we build on this realization to design a new hyperparameter-free unsupervised BM pipeline.
Our approach is inspired by two key insights:

\begin{enumerate}
    \item \textbf{Loss histories are rich representations.} Both JTT~\cite{liu_jtt_bias_mitigation_icml_21} and LfF~\cite{nam_lff_bias_mitigation_neurips_20} use a biased helper model to identify error slices for upweighting.
    Their high PoUs (seen in Figure~\ref{fig:model_selection}) suggest that when an identifier model learns a biased hypothesis, either by stopping at the ``right'' epoch (in JTT) or by properly calibrating a biasing loss (in LfF), it is a powerful way of identifying bias-conflicting samples.
    Nevertheless, as seen in Figure~\ref{fig:training_loss_curves}\textcolor{red}{a}, the fraction of conflicting samples captured by such a model quickly decreases as training progresses, and it does so at different rates for different tasks.
    This suggests that the discriminative power of a \textit{single} training ``snapshot'' is dataset-dependent and quickly lost once the model begins to overfit conflicting samples.
    
    Although an identifier's single training snapshot may fail to correctly identify bias-conflicting samples, we argue that the \textit{entire training history} of that model preserves a highly exploitable discriminative power.
    In Figure~\ref{fig:training_loss_curves}\textcolor{red}{b} we see that the mean training losses of bias-aligned and bias-conflicting samples represent two entirely different but salient \textit{clusters of signals}.
    This suggests that bias-conflicting and bias-aligned samples may be separated by splitting the training set into two groups, or \textit{clusters}, based on the samples' \textit{loss histories}.
    Assuming bias-conflicting samples represent \textit{a minority group within the population}, by looking at the smaller of the two discovered clusters we can then identify a training subset with a high representation of bias-conflicting samples.
    In Figure~\ref{fig:training_loss_curves}\textcolor{red}{c}, we see this for both \texttt{Waterbirds} and \texttt{CelebA},
    where loss-history clustering reveals a minority group with a notably higher representation of bias-conflicting samples compared to the original distribution.
    
    \begin{figure}[t!]
        \centering\includegraphics[width=\textwidth]{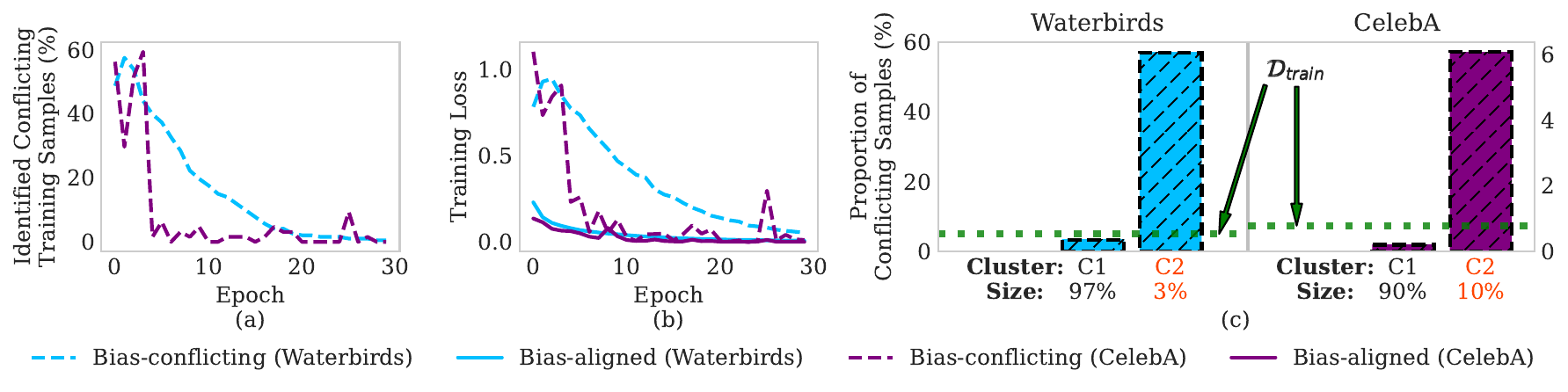}

        \caption{
            Training dynamics on \texttt{Waterbirds} (blue) and \texttt{CelebA} (purple).
            (a) Fraction of identified training set bias-conflicting samples.
            (b) Loss for bias-conflicting (dashed) and bias-aligned (solid) samples.
            (c) Proportion of bias-conflicting samples in the subsets discovered after clustering training samples based on their entire loss histories.
            In both tasks, the smaller cluster (red label) has a significantly higher proportion of bias-conflicting samples than the training distribution $\mathcal{D}_\text{train}$ (green line).
        }
        \label{fig:training_loss_curves}
    \end{figure}

    \item \textbf{Rebalancing enables hyperparameter-free upweighting.}
    Recent work has shown that group-balanced datasets encourage learning less biased models~\cite{kirichenko_last_layer_retraining_bias_mitigation_iclr_23, labonte_last_layer_retraining_neurips_23}.
    We observe that such robustness-inducing datasets can be constructed without any hyperparameters once a data subset with a higher representation of bias-conflicting samples has been identified.
    To see this, assume we partition $\mathcal{D}_\text{train}$ into two disjoint subsets, $\hat{\mathcal{D}}_\text{conflicting}$ and $\hat{\mathcal{D}}_\text{aligned}$, where $|\hat{\mathcal{D}}_\text{conflicting}| \leq |\hat{\mathcal{D}}_\text{aligned}|$ and $\hat{\mathcal{D}}_\text{conflicting}$'s proportion of bias-conflicting samples is larger than that of $\mathcal{D}_\text{train}$.
    With such a partition, we can construct a dataset $\mathcal{D}_\text{train}^\prime$ with a more balanced group distribution by first randomly selecting $\big(|\hat{\mathcal{D}}_\text{aligned}| - |\hat{\mathcal{D}}_\text{conflicting}|\big)$ elements from $\hat{\mathcal{D}}_\text{conflicting}$ to construct a multiset $\hat{\mathcal{D}}_\text{aug}$, and then joining this multiset with $\mathcal{D}_\text{train}$ to construct $\mathcal{D}_\text{train}^\prime$.
    This will make $\mathcal{D}_\text{train}^\prime$ have a higher proportion of bias-conflicting samples than $\mathcal{D}_\text{train}$, hence enabling access to a more group-balanced training set.
\end{enumerate}

\subsection{Proposed Pipeline}

Based on these insights, we introduce \textit{\textbf{T}argeted \textbf{A}ugmentations for \textbf{B}ias mitigation} (\textbf{TAB}). Our approach, summarised in Figure~\ref{fig:abstract} and in Algorithm~\ref{alg:tab} in Appendix~\ref{appendix:tab_algorithm}, is a hyperparameter-free unsupervised BM pipeline that exploits the \textit{entire} training history of a helper model to generate a group-balanced dataset for robust training.
Given a training set $\mathcal{D}_\text{train}$, a model $f_\theta$, and a learning algorithm $\mathcal{L}$ (e.g., ERM with SGD), TAB learns a robust classifier as follows:
\begin{enumerate}
    \item (\textbf{Loss history embedding construction}) First, we learn an identifier helper model $f_{\hat{\theta}_\text{base}}$ by running $\mathcal{L}$ on the training set $\mathcal{D}_\text{train} = \{(\mathbf{x}^{(i)}, y^{(i)})\}_{i = 1}^N$.
    While running $\mathcal{L}$, we keep track of the training loss of all samples in $\mathcal{D}_\text{train}$ throughout training.
    Hence, if we run $\mathcal{L}$ for $T$ steps, we construct a set of \textit{loss history embeddings} $H = \{\mathbf{h}^{(i)} \in \mathbb{R}^T\}_{i=1}^N$ where ${h}^{(i)}_t$ encodes the loss $\ell(f_{\hat{\theta}_\text{base}^{(t)}}(\mathbf{x}^{(i)}), y^{(i)})$ for sample $(\mathbf{x}^{(i)}, y^{(i)})$ at step $t$.

    \item (\textbf{Loss-aware partitioning})
    Once loss history embeddings are constructed, we exploit their discriminative power to separate bias-conflicting from bias-aligned samples (insight \#1).
    For this, we assign each sample in class $y \in \{1, \cdots, L\}$ one of two pseudo-groups discovered by splitting the history embeddings of samples in class $y$ into two clusters.
    Although several clustering methods would fit this purpose~\cite{ward1963hierarchical, ankerst1999optics, shi2003multiclass}, we cluster loss histories using $k$-means ($k = 2$) for simplicity.
    This partitions $\mathcal{D}_\text{train}$ into $2L$ ``pseudo-groups'' $\{(\mathcal{D}_l^{+}, \mathcal{D}_l^{-}) \; | \; l \in \{1, \cdots, L\}\}$, two per class, where, without loss of generality, we let $\mathcal{D}_l^{+}$ be the larger cluster for class $l$.
    We partition the data on a per-class basis since spurious correlations tend to be class-aligned.

    \item (\textbf{Group-balanced dataset generation}) Next, we generate a pseudo-group-balanced training set $\mathcal{D}_\text{train}^\prime$
    by upsampling each underrepresented subgroup, $\mathcal{D}_l^{-}$, to match the representation of samples in $\mathcal{D}_l^{+}$.
    Inspired by our insight~\#2 above, we achieve this as follows: for each class $l\in \{1, \cdots, L\}$, we construct an ``augmented'' multiset $\mathcal{D}_l^\prime$ by selecting $z_l = |\mathcal{D}_l^+| - |\mathcal{D}_l^-|$ samples from $\mathcal{D}_l^-$ uniformly at random.
    Then we construct a group-balanced dataset $\mathcal{D}_\text{train}^\prime$ by adding all augmentations to $\mathcal{D}_\text{train}$:
    \begin{equation}
        \mathcal{D}_\text{train}^\prime :=
            \Big( \bigcup_{l = 1}^L \mathcal{D}_l^+\Big) \cup \Big( \bigcup_{l = 1}^L \big(\mathcal{D}_l^- \cup \mathcal{D}_l^\prime \big)\Big) =
            \mathcal{D}_\text{train} \cup \Big( \bigcup_{l = 1}^L \mathcal{D}_l^\prime\Big)
    \end{equation}
    
    \item (\textbf{Robust model training}) Finally, we train a robust model $f_{\hat{\theta}_\text{TAB}}$ from scratch by running $\mathcal{L}$ on $\mathcal{D}_\text{train}^\prime$.
    As we expect $\mathcal{D}_\text{train}^\prime$ to be more group-balanced than $\mathcal{D}_\text{train}$, this results in $f_{\hat{\theta}_\text{TAB}}$ learning to generalise better.
\end{enumerate}


\section{Results}
\label{sec:results}
We study TAB via four questions capturing important practical considerations:

\begin{itemize}
    \item \textbf{(Q1) Worst-group Performance} -- Does TAB lead to better WGA compared to ERM in both controlled and real-world tasks?
    
    \item \textbf{(Q2.1) Average Performance} -- Do we observe a significant trade-off between mean accuracy and WGA in TAB models?

    \item \textbf{(Q2.2) Performance in Unbiased Datasets} -- In the absence of spurious correlations, does TAB maintain competitive average performance?

    \item \textbf{(Q3) Efficiency} -- What is the computational cost of using TAB as part of a model's pipeline? How does it compare to the costs of existing methods?

    \item \textbf{(Q4) Bias-conflicting Identification} -- Can TAB's loss history embeddings be used to identify spurious correlations in existing datasets?
\end{itemize}

\noindent We now describe our experimental setup
and then explore these questions.

\subsection{Setup}

\newheader{Datasets}
We explore a combination of synthetic and real-world vision tasks.
Our synthetic tasks are based on the Colour-MNIST~\cite{lit_mnist_dataset, nam_lff_bias_mitigation_neurips_20} dataset where samples are handwritten digits.
We construct two setups enabling evaluation on binary (i.e., $L = 2$) and multi-class (i.e., $L > 2$) tasks: (1) \texttt{Even-Odd}, whose task is to predict the digit's evenness ($L=2$), and (2) \texttt{cMNIST}, whose task is to predict the digit's value ($L=10$).
All digits are coloured with one of $L$ colours where there is a spurious correlation of strength $p \in [50\%, 100\%]$ between the $l$-th colour and the $l$-th class.
This is done by colouring $p\%$ of all images of label $l \in \{1, \cdots, L\}$ with the $l$-th colour, and the rest with a randomly selected colour.

We include three real-world datasets commonly used to evaluate model robustness: (1) \texttt{Waterbirds}~\cite{sagawa_gdro_bias_mitigation_19} ($L=2$), a binary bird classification where there is a spurious correlation between the background of the image and the class of the image;
(2) \texttt{CelebA}~\cite{liu_celeba_dataset} ($L=2$), a facial dataset where perceived gender is spuriously correlated with ``Blonde Hair'';
and,
(3) \texttt{BAR}~\cite{nam_lff_bias_mitigation_neurips_20} ($L=6$), an action dataset where each action is spuriously correlated with a scene.
Finally, we use \texttt{CUB}~\cite{wah_cub_dataset} ($L=200$), a bird classification task, as a control task without biases.

Although \texttt{BAR} has no group annotations, its test set has out-of-distribution background scenes for each action w.r.t.\ the train set.
Hence, for \texttt{BAR} we use worst-class accuracy as a proxy for the WGA.
Furthermore, to tractably explore different method hyperparameters within our computational budget (see below), for \texttt{CelebA} we use $15\%$ of the official training set ($N \approx 25,000$).
Finally, as \texttt{Waterbirds}'s standard validation and test sets are group-balanced, when computing mean accuracies we weight each sample based on their training group distributions to simulate in-distribution evaluation and avoid implicit privileged information leakage.
We include further details for all datasets in Appendix~\ref{appendix:datasets}.

\indent\newheader{Baselines}
We compare TAB against JTT~\cite{liu_jtt_bias_mitigation_icml_21} and LfF~\cite{nam_lff_bias_mitigation_neurips_20} due to their overlapping motivation with TAB.
We also include MaskTune~\cite{asgari_masktune_bias_mitigation_neurips_22}, a method that finetunes an ERM model after masking $\mathcal{D}_\text{train}$ using its saliency maps, as a recent popular unsupervised BM method.
Moreover, we include: (1) an ERM baseline as an upper-bound for mean accuracy and a lower-bound for WGA, and (2)~a G-DRO~\cite{hu_gdro_og} baseline as an upper-bound for WGA.
We emphasise that, with the exception of G-DRO, \textit{no method gets training or validation group information}.

\indent\newheader{Model Selection}
Given our observations in \S\ref{sec:importance_val_data}, we attempt to fairly evaluate all baselines via a \textit{thorough hyperparameter grid search} over a set of candidate hyperparameters for each method (chosen according to values used in their respective works).
We select models based on the hyperparameters with the \textit{best average validation accuracy} across three seeds.
All metrics we show are, therefore, averaged across three runs using the selected hyperparameters.
We avoid reusing hyperparameters from previous works to reduce leakage of group information through models selected using implicit or explicit group labels.

For all real-world tasks, we train a ResNet-18~\cite{resnet} whose initial weights are loaded from an ImageNet~\cite{deng_imagenet_dataset} pre-trained model.
In the synthetic tasks, we instead train a six-layer ReLU Convolutional Neural Network.
All models are trained using mini-batch stochastic gradient descent
and early stopping based on the validation accuracy.
Learning rates and $\ell_2$ decay factors across non-ERM methods are fixed based on values selected for ERM.
Finally, TAB's identifier model is trained as its ERM equivalent with the exception that early stopping is done based on the \textit{training accuracy} to interrupt training once the model converges.
All hyperparameters and training details are described in Appendix~\ref{appendix:hyperparameters}.

\subsection{Worst-group Performance (Q1)}

\begin{table}[t!]
    \setlength{\tabcolsep}{5pt}
    \centering
    \caption{
        Means and standard deviations of worst-group accuracy (WGA), price of unawareness (PoU), mean model selection WGA (MMS), and average accuracy.
        Models are selected using \textit{validation accuracy} over each method's shown hyperparameters.
        Note that (i) the set of hyperparameters $\Gamma$ for TAB is empty as it is hyperparameter-free, (ii) G-DRO utilises training group labels, and (iii) for \texttt{BAR} and \texttt{CUB} (tasks without group annotations), we use the worst-class accuracy as a proxy for WGA.
    }
    \label{table:metrics}

    \begin{adjustbox}{width=\textwidth, center}
    \begin{tabular}{cc||ccccc|c}
        \toprule
        {\parbox[t]{2mm}{\multirow{9}{*}{\rotatebox{90}{WGA (\%)}}}} & Method - \{Hypers\} &
             \texttt{Even-Odd} ($p=99\%$) &
             \texttt{cMNIST} ($p=98\%$) &
             \texttt{Waterbirds}&
             \texttt{CelebA} &
             \texttt{BAR} &
             \texttt{CUB} \\ \toprule
        {} & G-DRO - $\{\eta, \lambda_{\ell_2}\}$ &
            57.66 ± 6.76 &
            59.29 ± 3.27 &
            68.54 ± 1.75 &
            85.74 ± 0.69 & 
            N/A &
            N/A \\ \cmidrule(lr){2-8}
        {} & ERM - $\{\eta, \lambda_{\ell_2}\}$&
            55.98 ± 13.85  &
            46.97 ± 8.71  &
            44.86 ± 1.11   &
            34.81 ± 0.26 &
            29.56 ± 1.78 &
            16.67 ± 0.00  \\
        {} & LfF - $\{q\}$&
            2.97 ± 3.36 &
            48.45 ± 5.83 &
            51.14 ± 1.08 &
            \textbf{40.00 ± 0.00} &
            29.56 ± 2.35 &
            14.44 ± 3.14 \\
       {} & JTT - $\{T, \lambda_\text{up}\}$ &
            79.32 ± 1.76 &
            57.21 ± 3.59 &
            44.50 ± 0.45 &
            37.78 ± 2.83 &
            30.98 ± 2.00 &
            12.22 ± 1.57 \\
        {} & MaskTune - $\{\tau\}$ &
            72.82 ± 3.08 &
            13.94 ± 7.37 &
            35.67 ± 1.75 &
            37.04 ± 1.14 &
            17.61 ± 1.54 &
            10.00 ± 7.20 \\
        {} & TAB (ours) - $\emptyset$ &
            \textbf{81.85 ± 2.39} &
            \textbf{63.26 ± 2.50} &
            \textbf{55.92 ± 1.80} &
            \textbf{40.00 ± 1.20} &
            \textbf{38.94 ± 1.03} &
            \textbf{18.89 ± 1.57} \\ \toprule
        {\parbox[t]{2mm}{\multirow{4}{*}{\rotatebox{90}{PoU}}}} & ERM - $\{\eta, \lambda_{\ell_2}\}$&
            1.1181 ± 0.3195 &
            \textbf{1.0000 ± 0.0000} &
            1.0289 ± 0.0059 &
            1.0050 ± 0.0399 &
            \textbf{1.0000 ± 0.0000} & 
            N/A \\ 
        {} & LfF - $\{q\}$&
            5.2559 ± 0.0524 &
            1.2865 ± 0.2208 &
            1.3206 ± 0.0353 &
            1.6806 ± 0.0006 &
            1.3026 ± 0.1545 &
            N/A \\ 
       {} & JTT - $\{T, \lambda_\text{up}\}$&
            1.2462 ± 0.1054 &
            1.0805 ± 0.0807 &
            1.3588 ± 0.0571 &
            1.2475 ± 0.0253 &
            1.2321 ± 0.1149 & 
            N/A \\ 
        {} & MaskTune - $\{\tau\}$ &
            \textbf{1.0000 ± 0.0000} &
            1.9518 ± 0.2060 &
            1.5267 ± 0.0640 &
            1.0723 ± 0.0766 &
            2.3917 ± 0.6674 & 
            N/A \\ 
        & TAB (ours) - $\emptyset$ &
            \textbf{1.0000 ± 0.0000} &
            \textbf{1.0000 ± 0.0000} &
            \textbf{1.0000 ± 0.0000} &
            \textbf{1.0000 ± 0.0000} &
            \textbf{1.0000 ± 0.0000} & 
            N/A  \\\toprule 
        {\parbox[t]{2mm}{\multirow{4}{*}{\rotatebox{90}{MMS (\%)}}}} & ERM - $\{\eta, \lambda_{\ell_2}\}$&
            45.21 ± 3.39 &
            30.36 ± 1.71 &
            27.20 ± 2.59 &
            24.47 ± 2.36 &
            24.92 ± 2.92 &
            N/A \\ 
        {} & LfF - $\{q\}$&
            2.66 ± 0.79 &
            47.04 ± 8.79 &
            51.68 ± 0.69 &
            \textbf{45.56 ± 0.00} &
            32.53 ± 2.47 &
            N/A \\ 
       {} & JTT - $\{T, \lambda_\text{up}\}$&
            78.30 ± 1.13 &
            53.61 ± 4.94 &
            49.64 ± 0.30 &
            41.13 ± 0.55 &
            31.73 ± 2.75 &
            N/A \\
        {} & MaskTune - $\{\tau\}$ &
            39.05 ± 9.26 &
            6.59 ± 3.55 &
            46.77 ± 1.38 &
            29.89 ± 3.56 &
            17.10 ± 1.79 &
            N/A \\  
        & TAB (ours) - $\emptyset$ &
            \textbf{81.85 ± 2.39} &
            \textbf{63.26 ± 2.50} &
            \textbf{55.92 ± 1.80} &
            40.00 ± 1.20 &
            \textbf{38.94 ± 1.03} &
            N/A \\  \toprule
        {\parbox[t]{2mm}{\multirow{6}{*}{\rotatebox{90}{Mean Acc. (\%)}}}} & G-DRO - $\{\eta, \lambda_{\ell_2}\}$ &
            58.97 ± 6.79 &
            94.83 ± 0.55 &
            97.19 ± 0.28 &
            92.67 ± 0.14 &
            N/A &
            N/A \\ \cmidrule(lr){2-8}
        {} & ERM - $\{\eta, \lambda_{\ell_2}\}$&
            85.52 ± 12.09 &
            91.22 ± 0.26 &
            97.68 ± 0.06 &
            \textbf{95.45 ± 0.04} &
            56.93 ± 1.13 &
            \textbf{74.81 ± 0.29} \\
        {} & LfF - $\{q\}$&
            60.29 ± 13.53 &
            90.48 ± 1.17 &
            97.46 ± 0.12 &
            95.22 ± 0.02 &
            55.96 ± 1.25 &
            74.00 ± 0.67 \\
       {} & JTT - $\{T, \lambda_\text{up}\}$ &
            93.12 ± 4.74 &
            92.13 ± 1.13 &
            97.71 ± 0.11 &
            94.77 ± 0.05 &
            58.00 ± 2.34 &
            \camera{69.92 ± 0.10} \\
        {} & MaskTune - $\{\tau\}$ &
            92.60 ± 5.02 &
            83.25 ± 3.26 &
            \textbf{98.15 ± 0.04} &
            95.32 ± 0.07 &
            50.66 ± 1.38 &
            70.07 ± 0.97 \\
        {} & TAB (ours) - $\emptyset$ &
            \textbf{94.98 ± 3.37} &
            \textbf{93.28 ± 1.09} &
            97.52 ± 0.09 &
            94.67 ± 0.05 &
            \textbf{61.11 ± 0.94} &
            72.98 ± 0.34 \\ \bottomrule
    \end{tabular}
    \end{adjustbox}
\end{table}

\indent\newheader{TAB improves worst-group performance across all tasks.}
In Table~\ref{table:metrics} (top) we see that TAB achieves better WGA than the equivalent ERM baselines across all tasks.
This consistency in TAB's high WGA is not seen in competing methods:
we observe that (1)~LfF is highly unstable on \texttt{Even-Odd}, (2)~MaskTune can lead to worse WGA than ERM when its masking removes relevant features (e.g., \texttt{cMNIST} and \texttt{BAR}), and (3)~JTT does not improve ERM's WGA in \texttt{Waterbirds}.
Moreover, we observe that TAB outperforms all unsupervised baselines' WGAs in all tasks except for \texttt{CelebA}, where it performs on par with LfF.
Nevertheless,
TAB does not require a costly hyperparameter search or an unstable training objective,
something that cannot be said for LfF.
Finally, although G-DRO attains a higher WGA than TAB in most tasks as expected, we surprisingly observe that in highly-imbalanced tasks (e.g., \texttt{Even-Odd} and \texttt{cMNIST}), TAB surpasses G-DRO's WGA.
Together, all of these results suggest that TAB improves WGA across a variety of tasks and setups.

\indent\newheader{TAB avoids a hyperparameter search prone to selecting biased models.}
For all baselines, we perform a computationally expensive hyperparameter search on the hyperparameters shown next to each method's name.
The PoU computed for all methods (Table~\ref{table:metrics}, second quarter) suggests that competing baselines pay a considerable cost for having those hyperparameters: most of them can reach a high WGA, but such a hyperparameter configuration is discarded when performing accuracy-based model selection
(e.g., LfF in \texttt{Even-Odd} and JTT in \texttt{Waterbirds}).
More surprisingly, we observe that, on average,
competing BM pipelines are still prone to a worse WGA than TAB, as seen in the MMS score in Table~\ref{table:metrics} (third quarter) for \texttt{Even-Odd}, \texttt{cMNIST}, \texttt{Waterbirds}, and \texttt{BAR}.
This suggests that TAB, a method that requires no extra hyperparameter search, offers an effective BM alternative with good expected WGA.

\begin{figure}[t!]
    \centering
    \begin{subfigure}[b]{0.49\textwidth}
        \centering
        \includegraphics[width=0.8\textwidth]{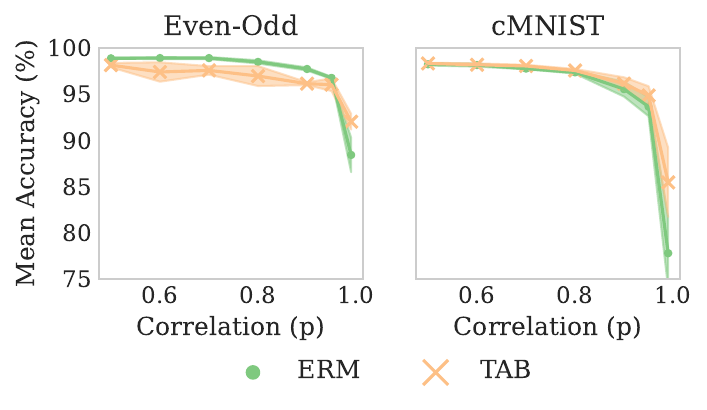}
    \end{subfigure}
    \begin{subfigure}[b]{0.49\textwidth}
        \centering
        \includegraphics[width=\textwidth]{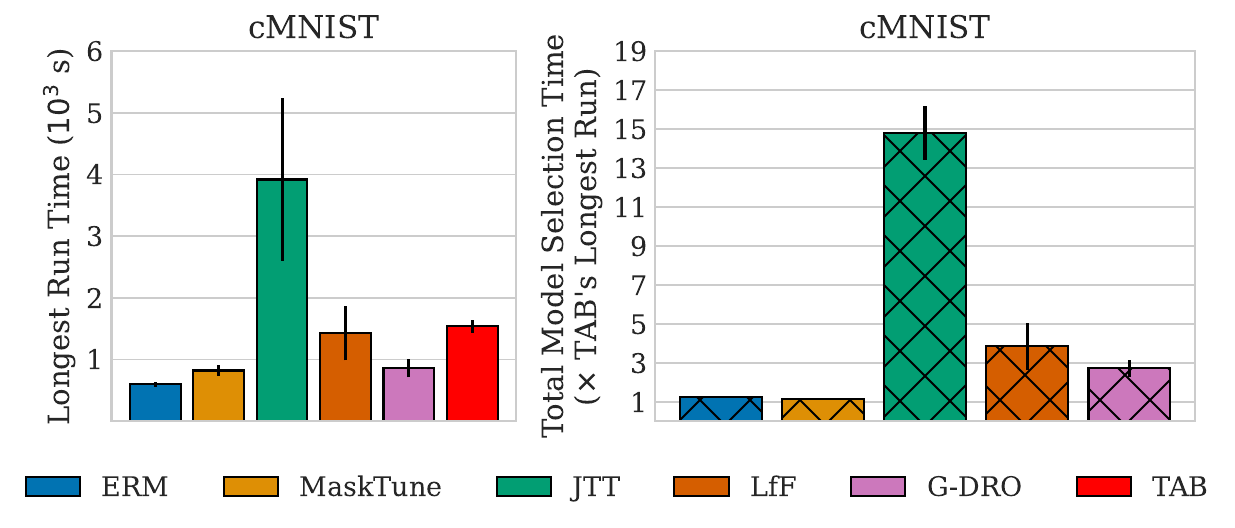}
    \end{subfigure}

    \caption{
        (Left) Mean accuracy for TAB and ERM
        as the correlation strength varies in \texttt{Even-Odd} and \texttt{cMNIST}.
        (Right) Worst run wall-clock time (solid bars) and total model selection time as a function of TAB's worst run (crossed bars) for all methods in \texttt{cMNIST}.
    }
    \label{fig:experiment_mixed_plots}
\end{figure}

\subsection{Average Performance (Q2)}
\label{sec:results_accuracy}

\indent\newheader{TAB achieves competitive mean accuracy against ERM models.}
In Table~\ref{table:metrics} (bottom), we show the average test accuracies for all baselines.
We observe that TAB achieves a competitive average accuracy w.r.t.\ ERM while maintaining a higher WGA across all tasks.
In particular, TAB attains the highest mean accuracies across all baselines for \texttt{Even-Odd}, \texttt{cMNIST},
and \texttt{BAR}, while suffering a small hit w.r.t.\ ERM for \texttt{Waterbirds} and \texttt{CelebA} (less than a 0.9\% absolute decrease).
These results suggest that TAB can simultaneously provide high mean accuracy and WGA across a variety tasks.

\indent\newheader{TAB's maintains high accuracy when biases disappear.}
We look at TAB's accuracy on three unbiased datasets.
First, in Figure~\ref{fig:experiment_mixed_plots} (left), we show TAB's and ERM's mean accuracies as the correlation strength $p$ in \texttt{Even-Odd} and \texttt{cMNIST} is ablated from $50\%$ (unbiased dataset) to $99.5\%$ (highly biased dataset).
Our results show that, in both instances, TAB achieves a mean accuracy similar to ERM's, even in completely unbiased setups.
Next, we explore TAB within \texttt{CUB}, a real-world task without any known biases.
As seen in the right-most column of Table~\ref{table:metrics},
TAB can not only maintain competitive mean accuracy in this unbiased dataset, but it can improve worst-class performance w.r.t.\ ERM.
This is in stark contrast with, say, JTT and MaskTune, where mean accuracy significantly deteriorates without any improvements to worst-class accuracy.
This suggests that, in contrast to competing approaches, TAB can be safely deployed in datasets where the existence of biases is unknown, a crucial use case for BM.

\subsection{Computational Efficiency (Q3)}
\label{sec:results_times}

\indent\newheader{Considering model selection, TAB's pipeline leads to shorter training times.}
In Figure~\ref{fig:experiment_mixed_plots} (right), we show the total wall-clock times taken to generate the results in Table~\ref{table:metrics} for the \texttt{cMNIST} task ($N \approx 40,000$).
We consider both the training time of the hyperparameters resulting in the longest run (left/solid) and the total time taken to train every attempted hyperparameterisation of each method as a function of TAB's single worst run (right/crossed).
Our results show that although a single TAB run can come with a computational penalty due to its identifier model (at its worst, about $2.5\times$ slower than ERM), this cost is significantly less than that of competing methods when considering their need for model selection (here, TAB is $1.14\times$ faster than MaskTune, the method with the second shortest total model-selection time, and $14.8\times$ faster than JTT, the method with the longest total time).
Overall, this makes TAB appealing for practical use in modern workloads where model selection scales poorly.

\subsection{Bias-conflicting Identification (Q4)}
\begin{table}[t!]
    \setlength{\tabcolsep}{5pt}
    \centering
    \caption{
        (Top) Representation of bias-conflicting (BC) examples in the original training set $\mathcal{D}_\text{train}$ and in TAB's augmented training set $\mathcal{D}_\text{train}^\prime$.
        (Bottom) Fraction of training BC examples correctly identified as underrepresented in TAB's loss history clustering.
    }

    \label{table:cluster_composition}
    \begin{adjustbox}{width=\textwidth, center}
    \begin{tabular}{c|cccc}
        {} &
             \texttt{Even-Odd} ($p = 99\%$) &
             \texttt{cMNIST} ($p = 98\%$) &
             \texttt{Waterbirds} &
             \texttt{CelebA} \\ \toprule
        BC Samples in $\mathcal{D}_\text{train}$ (baseline) &
            0.93\% ± 0.00\% &
            1.90\% ± 0.00\% &
            5.01\% ± 0.00\% &
            0.84\% ± 0.00\% \\
        BC Samples in $\mathcal{D}_\text{train}^\prime$ (after TAB's upsampling)&
            \textbf{62.64\% ± 0.39\%} &
            \textbf{56.79\% ± 1.83\%} &
            \textbf{66.47\% ± 3.16\%} &
            \textbf{9.39\% ± 0.63\%} \\ \hline
       Identified bias-conflicting samples &
            66.07\% ± 0.56\% &
            29.12\% ± 2.05\% &
            41.41\% ± 3.73\% &
            41.29\% ± 6.21\% \\ \bottomrule  
    \end{tabular}
    \end{adjustbox}
\end{table}

\indent\newheader{TAB augmentation leads to more group-balanced datasets.}
In Table~\ref{table:cluster_composition} (top) we show the representation of bias-conflicting examples in the original training sets of each task vs their representation in TAB's augmented dataset $\mathcal{D}_\text{train}^\prime$.
Across all tasks, we observe that TAB's splitting of the dataset based on loss history embeddings indeed leads to a significantly more group-balanced dataset: at its worst, TAB leads to an increase of $11.18\times$ in bias-conflicting sample representation (\texttt{CelebA}), while at its best, it leads to a $67.35\times$ increase in representation (\texttt{Even-Odd}).
Moreover, in Table~\ref{table:cluster_composition} (bottom) we see that TAB upsamples a significant portion of all bias-conflicting samples in the original training set (more than half for \texttt{Even-Odd}).
This suggests that TAB's upsampling leads to group-balanced datasets and may serve to inform future data collection.


\section{Discussion and Conclusion}
\label{sec:discussion}

\newheader{Limitations} \label{sec:discussion_limitations}
Although TAB improves WGA without group labels or model selection, it comes with some limitations.
First, \camera{the model training loop needs to be run twice}, which can become a limitation in \camera{massive datasets}.
Moreover, TAB's augmentations \camera{require storing the loss history matrix $H$ (which can have $\mathcal{O}(NT)$ size) and can lead to the training set doubling in size (in the worst case)}, increasing training times.
We argue, however, that these computational costs are amortised because TAB operates without a costly model selection.
A related limitation is that TAB requires $k$-means to operate, a method known to have scalability issues for large datasets.
Nevertheless, as shown in Appendix~\ref{appendix:batched_knn}, TAB's improvements can transfer when using a mini-batched $k$-Means~\cite{batch_kmeans}, a scalable clustering method.
Finally, we observe that in some highly imbalanced tasks, such as \texttt{CelebA}, TAB's WGA improvements over ERM are moderate compared to its improvements in other tasks (e.g., \texttt{Waterbirds}).
\camera{Future work could study how to improve this by (a)}~performing symmetry-invariant transformations of the augmented samples (we explore with some success in Appendix~\ref{appendix:transformation}), (b)~learning better discriminative embeddings through representations that capture both a sample's loss history and \textit{its semantics}, \camera{or (c)~exploiting additional information, such as the number of groups, by splitting the history embedding space into more than two clusters}.
\camera{For a further discussion on ethical considerations and potential negative societal impacts of this work, please refer to Appendix~\ref{sec:discussion_ethics}}.


\indent\newheader{Conclusion}
\label{sec:conclusion}
In this paper, we studied biases in the form of spurious correlations and their effects on DNNs.
We first argued that current unsupervised bias-mitigation methods are prone to learning biased models due to the inherent difficulty of group-unsupervised model selection.
Then, we addressed this gap by proposing Targeted Augmentations for Bias mitigation (TAB), a simple but powerful hyperparameter-free bias mitigation pipeline that can be used on any learning pipeline.
TAB learns a robust classifier by first discovering bias-conflicting training samples based on the entire loss histories of a pre-trained model, and then generating a group-balanced dataset from which a new model can be trained.
We showed that TAB significantly increases worst-group accuracy across multiple tasks without any model selection or a severe sacrifice to mean accuracy.
These results suggest that TAB, an approach requiring only a few lines of code to be implemented,
represents a move towards \textit{practical debiasing}.

\newpage

\section*{\camera{Acknowledgements}}
\camera{This work was funded by Sony Research. The authors would like to thank Apostolos Modas, William Thong, Wiebke Hutiri, and Andrei Margeloiu for their insightful comments and feedback on previous iterations of this manuscript. MEZ acknowledges further support from the Gates Cambridge Trust via a Gates Cambridge Scholarship.}

%
%
\bibliographystyle{splncs04}
\bibliography{main}

\newpage
\renewcommand\thefigure{\thesection.\arabic{figure}}
\setcounter{figure}{0}
\renewcommand\thetable{\thesection.\arabic{table}}
\setcounter{table}{0}
\renewcommand\theequation{\thesection.\arabic{equation}}
\setcounter{equation}{0}
\renewcommand\thealgorithm{\thesection.\arabic{algorithm}}
\setcounter{algorithm}{0}

\appendix

\section{\camera{Ethical Considerations}}
\label{sec:discussion_ethics}

\camera{Our work improves a DNN's fairness through a pragmatic and effective pipeline.
We hope this may lead to TAB being included as a standard debiasing good practice in real-world workloads.
Nevertheless, we recognise there are potential negative societal aspects to consider.
First, TAB trains its model twice, doubling the carbon footprint of training~\cite{anthony2020carbontracker}.
This, however, can be amortised in practice by TAB's lack of model selection.
Second, the ability of TAB to identify a large portion of bias-conflicting samples implies that bad actors may use this information to further increase a model's bias (e.g., by removing these training samples).
Nevertheless, we argue that this ill-intended use of our method is not unique to our approach, but rather an inherent but unlikely negative aspect of all BM methods.}

\section{Summary of Previous Bias Mitigation Methods}
\label{appendix:previous_work}
To complement our related work discussion in Section~2, in this section we include a summary of previous work that is closely aligned with our proposed method.
In Table~\ref{table:previous_bias_mitigation}, we show several bias mitigation methods together with (1) their level of group supervision (whether they need group labels in the training set, in the validation set, or in both), and (2) the set of key hyperparameters that require fine-tuning each method.
We note, however, that most of these approaches have more underlying hyperparameters than those shown as they may still require fine-tuning of hyperparameters specific to the target model (e.g., the underlying model's weight decay, learning rate, batch size, etc).
Nevertheless, as these are hyperparameters that are not unique to any BM method but rather a necessity of the model chosen to be debiased, in Table~\ref{table:previous_bias_mitigation} we show only the BM-specific hyperparameters.
By contrasting methods this way, we see how TAB is unique within fully unsupervised BM approaches by not requiring any extra hyperparameters, avoiding a potentially costly model selection and enabling easy adoption in modern training pipelines.

\begin{table}[h!]
    \setlength{\tabcolsep}{5pt}
    \centering
    \caption{
        Bias mitigation methods together with their group annotation requirements.
        To highlight the practical considerations involved when using any of these approaches, we include the number and identity of key sensitive hyperparameters for each approach. However, we note that these are hyperparameters required \textbf{on top of the standard ERM hyperparameters needed to train a model} (e.g., ``learning rate'', ``batch size'', etc.).
        \textsuperscript{\dag} These papers were published around the time of our submission and, therefore, constitute concurrent works.
    }
    \label{table:previous_bias_mitigation}
    \begin{adjustbox}{width=\textwidth, center}
    \begin{tabular}{c|ccc}
        \toprule
        Method &
            Train Groups &
            Val Groups &
            \# of Hyperparameters - \{Hypers\}\\ \midrule
            
        {Adversarial Debiasing~\cite{zhang_adversarial_bias_mitigation_18}} &
            \cmark &
            \cmark &
            2 - \{$A$, $\alpha$\} \\
        {G-DRO~\cite{sagawa_gdro_bias_mitigation_19}} &
            \cmark &
            \cmark &
            4 - \{$C$, $\lambda$, $\eta_q$, $\eta_\theta$\}\\
        {PGI~\cite{ahmed_pgi_bias_mitigation_icml_20}} &
            \cmark &
            \cmark &
            1 - \{invariance penalty\}\\
        {DFR~\cite{kirichenko_last_layer_retraining_bias_mitigation_iclr_23}} &
            \cmark &
            \cmark &
            2 - \{retrain epochs, \# retrains\} \\ \hline
            
        {Multiaccuracy~\cite{kim_multiaccuracy_bias_mitigation_19}} &
            \xmark &
            \cmark &
            2 - \{$\mathcal{A}$, $\alpha$\}\\
        {CVaR DRO~\cite{levycvar_dro_bias_mitigation_neurips_20}} &
            \xmark &
            \cmark &
            1 - \{$\alpha$\}\\
        {LfF~\cite{nam_lff_bias_mitigation_neurips_20}} &
            \xmark &
            \cmark &
            1 ($q$) \\
        {JTT~\cite{liu_jtt_bias_mitigation_icml_21}} &
            \xmark &
            \cmark &
            2 - \{$T$, $\lambda_\text{up}$\}\\
        {Spectral Decoupling~\cite{pezeshki_gradient_starvation_bias_mitigation_neurips_21}} &
            \xmark &
            \cmark &
            2 - \{$\lambda$, $\gamma$\}\\
        {EIIL~\cite{creager_eiil_icml_21}} &
            \xmark &
            \cmark &
            1 or more - \{underlying learner dependent\}\\
        {CIM~\cite{taghanaki_cim_bias_mitigation_icml_21}} &
            \xmark &
            \cmark &
            1 - \{$\alpha$\}\\
        {OccamNet~\cite{shrestha_occamnets_bias_mitigation_eccv_22}} &
            \xmark &
            \cmark &
            2 - \{$\tau_{\text{acc}, 0}$, $\lambda_\text{CS}$\}\\
        {SELF~\cite{labonte_last_layer_retraining_neurips_23}} &
            \xmark &
            \cmark &
            2 or more - \{$n$, $g$\}\\
            \hline

        {GEORGE~\cite{sohoni_george_bias_mitigation_neurips_20}} &
            \xmark &
            \xmark &
            4 or more - \{$f_\theta$, proj. comps, $F$, $s_\text{min}$\}\\
        {MaskTune~\cite{asgari_masktune_bias_mitigation_neurips_22}} &
            \xmark &
            \xmark &
            2 or more - \{$\tau$, $\mathcal{G}$ \}\\
        {DebiAN~\cite{li_debian_bias_mitigation_eccv_22}} &
            \xmark &
            \xmark &
            1 or more - \{discoverer $D$\}\\
        {AGRO~\cite{paranjape_agro_bias_mitigation_iclr_23}} &
            \xmark &
            \xmark &
            6 or more - \{$T_1$, $T_2$, $m$, $\alpha$, $\lambda$, $\phi_h$\}\\
        {CB Last-layer retraining\textsuperscript{\dag}~\cite{labonte_last_layer_retraining_neurips_23}} &
            \xmark &
            \xmark &
            1 - \{retrain epochs\}\\
        {uLA\textsuperscript{\dag}~\cite{tsirigotis_no_labels_ssl_bias_mitigation_neurips_23}} &
            \xmark &
            \xmark &
            4 - \{$T_\text{ssl}$, $T_\text{stop}$, $\eta$, $\tau$\} \\ \hline
            
        \textbf{TAB (ours)} &
            \xmark &
            \xmark &
            0 - $\emptyset$ \\ \bottomrule       
    \end{tabular}
    \end{adjustbox}
\end{table}

\section{Price of Unawareness and Mean Model Selection}
\label{appendix:pou}

In this section, we formally define the metrics discussed in Section~4 and describe how we can compute such metrics in practice.

\indent \newheader{Price of Unawareness (PoU)}
Borrowing inspiration from similar concepts in game theory~\cite{koutsoupias1999worst, papadimitriou2001algorithms}, we capture the cost ``paid'' by a BM pipeline when lacking proper group labels during model selection through its \textit{price of unawareness}.
Formally, given a BM pipeline $\mathcal{B}_\gamma$ with hyperparameters $\gamma \in \Gamma$,
let $\hat{\theta}(\mathcal{B}_\gamma, \mathcal{P}_\text{train})$ be the parameters $\hat{\theta} \in \Theta$ learnt by running $\mathcal{B}_\gamma$ on a training set formed by $N$ i.i.d. samples from $\mathcal{P}_\text{train}$.
In this setup, we define the PoU of $\mathcal{B}$ as follows:
\begin{align}
    \label{eq:pou}
    \text{PoU}(\mathcal{B}, \mathcal{P}_\text{train}, \mathcal{P}_\text{test}, \mathcal{P}_\text{val}) := \frac{\max_{\gamma^\prime \in \Gamma} \text{WGA}\big(f_{\hat{\theta}(\mathcal{B}_{\gamma^\prime}, \mathcal{P}_\text{train})}, \mathcal{P}_\text{test}\big)}{\text{WGA}\big(f_{\hat{\theta}(\mathcal{B}_{\gamma_\text{val-acc}}, \mathcal{P}_\text{train})}, \mathcal{P}_\text{test} \big)}
\end{align}
where $\mathcal{P}_\text{train}$, $\mathcal{P}_\text{test}$, and $\mathcal{P}_\text{val}$ are the training, testing, and validation distributions (assumed to be the same but written separately for clarity), and $\gamma_\text{val-acc}$ are the hyperparameters chosen when performing model selection based on the validation average accuracy (for notational clarity, we do not explicitly write their dependency on $\mathcal{B}$, $\mathcal{P}_\text{train}$, and $\mathcal{P}_\text{val}$).
That is:
\begin{equation}
\label{eq:hypers_val_acc}
    \gamma_\text{val-acc} := \argmax_{\gamma \in \Gamma}  \mathbb{E}_{(\mathbf{x}, y) \sim \mathcal{P}_\text{val}(\mathbf{x}, y)}\big[ \mathbbm{1}\big(f_{\hat{\theta}(\mathcal{B}_\gamma, \mathcal{P}_\text{train})}(\mathbf{x}) = y\big) \big]
\end{equation}

\noindent In practice, we estimate the PoU through a discrete grid search over $\Gamma_\text{cands} \subseteq \Gamma$ while ensuring that samples in the training, testing, and validation sets are entirely disjoint.
If during this grid search we find that the test WGA of $f_{\hat{\theta}(\mathcal{B}_{\gamma_\text{val-acc}}, \mathcal{P})}$ is $0$ for some hyperparameters $\gamma$, then we discard these hyperparameters as otherwise the PoU may become infinite and non-informative.
Hence, we estimate this value based on hyperparameterisations with non-zero validation WGA only.

We note that, if the validation distribution $\mathcal{P}_\text{val}$ is the same as the testing distribution $\mathcal{P}_\text{test}$, as it is commonly the case for in-distribution domains, then for large enough validation sets (i.e., $N >> 1$), the PoU is equivalent to computing the ratio between the WGA of the model selected using \textit{validation WGA} and the WGA of the model selected using validation mean accuracy.
Nevertheless, because access to a large validation set is rare, we defined the numerator of equation~\ref{eq:pou} in terms of the test distribution to provide a realistic upper bound for a method's achievable WGA.
This enables us to compute a PoU that (1) is stable even in extremely group-imbalanced domains where minority groups may be missing in a small validation set, and (2) is guaranteed to be bounded below by 1, enabling easy interpretation of the PoU as discussed in Section~4.

\indent \newheader{Mean Model Selection WGA (MMS)}
In contrast to the PoU, which aims to capture the worst-case cost of lacking validation group labels during BM, the Mean Model Selection WGA (MMS) is designed to capture the expected behaviour of a BM pipeline across multiple hyperparameters.
In other words, we are interested in measuring the expected WGA across all possible hyperparameters.
This serves as a proxy measurement of what one expects to get from a BM approach when deploying it across a task without an extensive model selection.
The MMS is therefore formally defined as follows:
\begin{align}
    \label{eq:mms_formal}
    \text{MMS}(\mathcal{B}, \mathcal{P}_\text{train}, \mathcal{P}_\text{test}) := \mathbb{E}_{\gamma^\prime \sim \text{Unif}(\Gamma)} \Big[\text{WGA}\big(f_{\hat{\theta}(\mathcal{B}_{\gamma^\prime}, \mathcal{P}_\text{train})}, \mathcal{P}_\text{test}\big) \Big]
\end{align}
where, as in the case for our fomarlisation of the PoU, $\mathcal{B}_\gamma$ is a BM pipeline taking hyperaparameters $\gamma \in \Gamma$.

In practice, we follow a similar approach as done with the PoU and estimate the MMS using a grid search over a set of potential hyperparameters $\Gamma_\text{cands} \subseteq \Gamma$.
More specifically, we compute this estimate using hyperparameters in $\Gamma_\text{cands}$ to compute an empirical Monte Carlo estimate of the MMS:
\begin{align}
    \label{eq:mms_practice}
    \text{MMS}(\mathcal{B}, \mathcal{P}_\text{train}, \mathcal{P}_\text{test}) \approx \frac{1}{|\Gamma_\text{cands}|} \sum_{\gamma^\prime \in \Gamma_\text{cands}} \text{WGA}\big(f_{\hat{\theta}(\mathcal{B}_{\gamma^\prime}, \mathcal{P}_\text{train})}, \mathcal{P}_\text{test}\big)
\end{align}

\section{Targeted Augmentations for Bias Mitigation}
\label{appendix:tab_algorithm}

In Section~5 we provide a detailed motivation and description of TAB, our proposed unsupervised bias mitigation pipeline.
Here, we complement that discussion by providing pseudocode for TAB (Algorithm~\ref{alg:tab}).
This format highlights two key things: first, as discussed in Section~5, TAB takes no extra hyperparameters on top of those of its underlying learning algorithm $\mathcal{L}$.
Second, TAB's implementation is extremely simple and can be easily added to existing training pipelines.
We believe that both of these key properties make TAB a likely candidate for adoption in practical real-world scenarios.

\begin{algorithm}[h!]
   \caption{TAB}
   \label{alg:tab}
    \begin{algorithmic}
        \State {\bfseries Input:}  Learning algorithm $\mathcal{L}$, target neural network $f_\theta$,
        and
        training set $\mathcal{D}_\text{train} = \{(\mathbf{x}^{(i)}, y^{(i)}) \; | \; \mathbf{x}^{(i)} \in \mathbb{R}^n, y^{(i)} \in \big\{1, \cdots, L\} \big\}_{i = 1}^N$
        \State {\bfseries Output:} Robust model $f_{\hat{\theta}_\text{TAB}}$
        
        \State $f_{\hat{\theta}_\text{base}}$, $\{ \mathbf{h}^{(i)} \in \mathbb{R}^T \}_{i = 1}^N$ $\leftarrow \mathcal{L}\big(f_\theta, \mathcal{D}_\text{train})$ \Comment{Train identifier and build loss histories $\mathbf{h}^{(i)}$}
        \For{$l \in \{1, \cdots, L\}$}
            \State $\mathcal{H}_l \leftarrow \{\mathbf{h}^{(i)} \; | \; y^{(i)} = l\}$ \Comment{Get loss histories of all samples with label $l$}
            \State $\mathcal{D}_l^+, \mathcal{D}_l^- \leftarrow \text{k-means}\big(\mathcal{H}_l, k=2)$ \Comment{Split all samples into two clusters}
            \If{$|\mathcal{D}_l^+| < |\mathcal{D}_l^-|$}
                \State $\mathcal{D}_l^+, \mathcal{D}_l^- \leftarrow \mathcal{D}_l^-, \mathcal{D}_l^+$ \Comment{$\mathcal{D}_l^-$ will be the minority cluster}
            \EndIf
            \State $\mathcal{D}_{l}^\prime \leftarrow \emptyset$ \Comment{Construct this class' augmented \textbf{multiset}}
            \While{$|\mathcal{D}_{l}^\prime| < |\mathcal{D}_{l}^+| - |\mathcal{D}_{l}^-|$}
                \State $j \leftarrow \text{UniformSample}\big( \mathcal{D}_l^- \big)$ \Comment{Randomly select a sample index from $\mathcal{D}_l^-$}
                \State $\mathcal{D}_{l}^\prime \leftarrow \mathcal{D}_{l}^\prime \cup \{ (\mathbf{x}^{(j)}, y^{(j)})\}$ \Comment{Add the $j$-th training sample to $\mathcal{D}_{l}^\prime$}
            \EndWhile
        \EndFor
        \State $\mathcal{D}_{\text{train}}^\prime \leftarrow \mathcal{D}_\text{train} \cup \Big(\bigcup_{l = 1}^L \mathcal{D}_l^\prime \Big)$ \Comment{Construct our augmented training \textbf{multiset}}
        
        \State $f_{\hat{\theta}_\text{TAB}}$ $\leftarrow \mathcal{L}\big(f_\theta, \mathcal{D}_\text{train}^\prime\big)$ \Comment{Learn robust model from scratch}
        \State {\bfseries Return:} $f_{\hat{\theta}_\text{TAB}}$
    \end{algorithmic}
\end{algorithm}

\section{Dataset Details}
\label{appendix:datasets}
In this section, we provide a detailed description of the datasets we use for our evaluation in Section~6.
Specifically, we show the main characteristics of each task in Table~\ref{table:datasets} and describe each of the specific tasks in more detail below.

\begin{table}[!htbp]
    \centering
    \caption{
        Key characteristics of all datasets used in this paper.
        All tasks in our evaluation are classification tasks with $L$ categorical labels.
        We show the total number of groups $k$ for each task as well as its \textit{worst-group size}, defined as the ratio between the number of samples in the smallest group and the number of samples in the entire dataset.
        Both \texttt{BAR} and \texttt{CUB} have not known group labels.
        Nevertheless, for \texttt{BAR} there is a distribution shift between the training and the test set where actions are displayed on different backgrounds than those used in the training set.
    }
    \label{table:datasets}
    \begin{adjustbox}{width=\textwidth, center}
    \begin{tabular}{c|ccccc}
        {Dataset} &
            \# of Samples ($N$) &
            \# of Features ($n$) &
            \# of Classes ($L$) &
            \# of Groups ($k$) & 
            Worst-Group Size (\%) \\ \toprule
        \texttt{Even-Odd} ($p = 99\%$)& 48,000 & (3, 28, 28) & 2 & 4 & 0.41\% \\
        \texttt{cMNIST} ($p = 98\%$)& 48,000 & (3, 28, 28) & 10 & 100 & 0.01\% \\
        \texttt{Waterbirds} & 4,795 & (3, 224, 224) & 2 & 4 & 1.17\% \\
        \texttt{CelebA} (subsampled) & 24,416 & (3, 224, 224) & 2 & 4 & 0.83\% \\ \hline
        \texttt{BAR} & 1,552 & (3, 224, 224) & 6 & 12 & Unknown \\
        \texttt{CUB} & 4,796 & (3, 224, 224) & 200 & Unknown & Unknown \\
        \bottomrule
    \end{tabular}
    \end{adjustbox}
\end{table}


\indent\newheader{Even-Odd}
The \texttt{Even-Odd} task is a synthetic binary ($L=2$) visual classification task where the strength of existing spurious correlations can be easily controlled.
Taking inspiration from the Colour-MNIST~\cite{nam_lff_bias_mitigation_neurips_20} dataset, we construct our task by colouring handwritten digits from the \texttt{MNIST}~\cite{lit_mnist_dataset} dataset with two colours.
Specifically, each sample $\mathbf{x}^{(i)}$ of the original MNIST task is transformed from a $28 \times 28$ grayscale image to a $3 \times 28 \times 28$ normalised RGB image (i.e., $\mathbf{x}^{(i)} \in [0, 1]^{3 \times 28 \times 28}$) by colouring the digit with either ``red'' or ``green'' hues proportional to the grayscale pixel intensities.
We define a binary task where the label determines whether an image is ``odd'' ($y = 0$) or ``even'' ($y = 1)$ and introduce a spurious correlation of strength $p\in [50\%, 100\%]$ by selecting $p\%$ of all ``odd'' samples and colouring them ``red'' while colouring the rest of $1 - p$ samples using ``green''.
Similarly, we introduce an equivalent correlation with strength $p$ between ``even'' and ``green'' samples.
This yields a task with $4$ groups, one for each pair $(\text{label}, \text{colour})$, of which the smallest subgroup has an expected size of $N \times \frac{1}{2} \times \frac{100 - p}{100}$.

The training, testing, and validation sets in \texttt{Even-Odd} are constructed using the procedure above on the standard train, test, and validation sets in the MNIST task.
We note that to enable accurate evaluation, we use a balanced test set that has an equal representation on all $(\text{class}, \text{colour})$ combinations.
This allows us to accurately estimate the WGA, as otherwise the smallest group in the test set is too small for getting a good estimate of the model's accuracy on members of that group.
Nevertheless, when computing test mean accuracy on this group-balanced test dataset, we weight each sample based on the representation of their respective $(\text{class}, \text{colour})$ group in the training distribution.
That way, the mean average accuracy in the test set behaves as if it is from the same distribution as the training set.

\indent\newheader{cMNIST}
Similar to \texttt{Even-Odd}, the \texttt{cMINST} task is a Colour MNIST-based classification task with a predefined spurious correlation strength.
Each sample $\mathbf{x}^{(i)}$ of the original MNIST dataset is transformed from a $28 \times 28$ grayscale image to a $3 \times 28 \times 28$ normalised RGB image (i.e., $\mathbf{x}^{(i)} \in [0, 1]^{3 \times 28 \times 28}$) by colouring the digit with one of 10 colours (the colours are randomly-generated RGB colours fixed beforehand).
In this dataset, we define a multilabel task  ($L = 10$) where the label determines the digit's identity ($y \in \{0, 1, \cdots, 9\}$.
As in \texttt{Even-Odd}, we introduce a spurious correlation of strength $p\in [50\%, 100\%]$ by selecting $p\%$ of all samples with any given class label $l\in \{0, 1, \cdots, 9\}$ and colouring them the $l$-th colour while colouring the rest of $1 - p$ samples of that class with a colour randomly selected from the remaining $L -1$ colours.
This results in a task with $L^2$ groups, one for each pair $(\text{label}, \text{colour})$ of which the smallest subgroup has an expected size of $N \times \frac{1}{10} \times \frac{100 - p}{100}$.
We generate the training, testing, and validation sets as in `\texttt{Even-Odd}'', constructing a group-balanced test set for stability in evaluation and using a weighted accuracy to compute the mean accuracy.

\indent\newheader{Waterbirds}
The \texttt{Waterbirds}~\cite{sagawa_gdro_bias_mitigation_19} task is a binary bird classification task ($L = 2$) where each sample is formed by an image of a bird (selected from the CUB dataset~\cite{wah_cub_dataset}) on top of a either a ``land`` background or a ``water'' background (selected from the Places dataset~\cite{zhou2017places}).
Birds in this task are split between ``waterbirds'' (birds that are either seabirds or waterfowl) and ``landbirds'' (rest of birds).
There is a spurious correlation introduced between the type of bird (i.e., the downstream label $y$) and the background selected from the Places dataset.
Specifically, images from the ``ocean'' and ``natural lake'' categories in Places are spuriously correlated with ``waterbirds'' (i.e., $y = 0$), while backgrounds from the ``bamboo'' or ``broadleaf'' forest categories are spuriously correlated with ``landbirds'' ($y = 1$).
This results in a total of $4$ subgroups $\{$(``waterbird'', ``water background''), (``waterbird'', ``land background''), (``landbird'', ``water background''), (``landbird'', ``land background'')$\}$ of which (``waterbird'', ``land background'') and (``landbird'', ``water background'') are minority bias-conflicting subgroups.

We use the same training/validation/testing splits
from Sagawa et al.~\cite{sagawa_gdro_bias_mitigation_19} however during validation and testing, as in the MNIST-based tasks, we use a weighted accuracy for the mean accuracy where every sample is weighted based on the training distribution of its (label, background) subgroup in the training set.
This is because, in the standard \texttt{Waterbird} splits, the training set is highly group-imbalanced while the validation and test sets are both group-balanced.
Therefore, using the validation set as is may result in accidental group information being leaked during model selection (an unrealistic assumption as the validation set is usually sampled from the same distribution as the training set).
All images are resized to be $3\time 224 \times 224$ RGB images (in floating point representation) and are normalised using the mean and standard deviations of the ImageNet~\cite{deng_imagenet_dataset} dataset.
Finally, during training we perform random croppings and horizontal flips.

\indent\newheader{CelebA}
We construct a real-world human-centric task based on the CelebA face recognition dataset~\cite{liu_celeba_dataset}, a large collection of celebrities' face images collected from the internet annotated with an identity label and a set of 40 binary attributes.
We follow the approach by Sagawa et al.~\cite{sagawa_gdro_bias_mitigation_19} and construct a binary task ($L = 2$) out of this dataset by using the ``Blonde Hair'' attribute as the downstream label.
We use this setup as there is a strong spurious correlation between the annotated perceived gender of each image and the annotated colour of their hair.
In this setup, samples annotated as ``not blonde'' and ``male'' are significantly underrepresented (worst bias-conflicting group), leading to models exploiting perceived gender to determine the hair colour.
More generally, the bias-conflicting samples correspond to (``blonde'', ``male'') and (``not blond'' and ``female'') samples,
with the (``blonde'', ``male'') group being significantly smaller than the
(``not blonde'', ``female'') group.
Although the original splits of the CelebA dataset contain $162,770$ training samples and $19,867$ test samples, for our experiments we use $15\%$ of the training set to enable a tractable and exhaustive model selection across all methods with hyperparameters as discussed above (this is particularly important for costly methods such as JTT).
This results in a training set with approximately $24,000$ samples.
Furthermore, we also use $15\%$ of the validation set for consistency and resize all samples to $3\times 224 \times 224$ images while applying random horizontal flips during training.
All images are turned into their floating-point representation and normalised using ImageNet~\cite{deng_imagenet_dataset}'s means and standard deviations as normally done for this task.

\indent\newheader{BAR}
Next, we evaluate all models on the Biased Action Recognition (\texttt{BAR}) task, a real-world action detection dataset.
This dataset contains samples of humans performing one of six actions (``Climbing'', ``Fishing'', ``Racing'', ``Throwing'', ``Vaulting''), and we are tasked with predicting each action from the image ($L=6$).
Each action in this dataset is spuriously correlated with a specific scene: ``Climbing'' actions are most commonly done on ``Rock Walls'', ``Diving'' actions are most commonly done ``Underwater'', ``Fishing'' actions are most commonly done on the ``Water Surface'', ``Racing'' actions are most commonly done on a ``Paved Track'', ``Throwing'' actions are most commonly done on a ``Playing Field'', and ``Vaulting'' actions are most commonly shown in front of ``Sky'' backgrounds.
This dataset does not have group annotations, making it impossible to accurately compute the WGA during training, testing, or validation.
Nevertheless, the test set is constructed so that actions are more commonly shown with scenes that are not aligned with the action's spurious background found in the training set.
Therefore, the mean accuracy and worst-class accuracy in the test set serve as a good proxy for WGA for this task, given the distribution shift between the train and test sets.

We use the original train and test splits.
However, we randomly select $20\%$ of the training set as a validation set.
All images are resized to be $3\time 224 \times 224$ RGB images (in floating point representation) and are normalised using the mean and standard deviations of the ImageNet~\cite{deng_imagenet_dataset} dataset.
Finally, during training we perform random scalings and horizontal flips.

\indent\newheader{CUB}
Finally, we use the \texttt{CUB} dataset as a real-world task without known spurious correlations.
Samples in this task correspond to RGB images of birds and are annotated with their bird type identity ($L = 200$) as the downstream label.
We process all images as in~\cite{cbm} by normalising and randomly flipping and cropping each image during training (normalisation is done using ImageNet's mean and standard deviation).
This results in a dataset with approximately $6,000$ images with shape $3 \times 299 \times 299$ that is split into a train, validation, and test subsets using the same splits as Koh et al.~\cite{cbm} (training set has approximately $4,800$ images).

\section{Experimental Details}
\label{appendix:hyperparameters}
In this section, we describe our experimental setup across all tasks and baselines for our experiments described in Section~6.
We begin by describing the underlying architecture and optimisation setup used across baselines.
Then, we proceed to describe each method's hyperparameter selection process.

\subsection{Architecture and Optimisation Setups}
\label{sec:appendix_training_hyperparameters}

For all MNIST-based synthetic tasks (i.e., \texttt{Even-Odd} and \texttt{cMNIST}), we train as the underlying model of all baselines a six-layered Convolutional Neural Network (CNN).
Specifically, we construct a CNN formed by six 2D Convolutional layers with $3\times 3$ filters, ``same'' padding, and $\{16, 16, 32, 32, 64, 64\}$ output feature maps.
A ReLU non-linear activation follows each of these layers and a single linear layer is used to map the output of the last convolution to the number of output classes (no activation is used after this layer as we work with logit outputs).
We learn this model's parameters via an Adam optimiser~\cite{kingma2014adam} with its default configuration (learning rate $\eta = 10^{-3}$, $\beta_1 =0.9$, $\beta_2 = 0.999$). Moreover, given the smaller model sizes for these tasks, we use a relatively large batch size of $2,048$ samples to better utilise our hardware.
Notice that in these synthetic tasks, we do not perform a hyperparameter search on the optimiser's learning rate as we observed good performance with the default optimiser configuration and a significant drop in ERM validation accuracy when this learning rate was decreased.

In contrast, for our real-world tasks we use a Resnet-18~\cite{resnet} architecture whose weights for all layers but the output linear layer are initialised to those from a Resnet-18 model pretrained on ImageNet~\cite{deng_imagenet_dataset} (using Pytorch Vision's default loaded weights~\cite{pytorch}).
To reduce memory consumption and parallelise model selection, we use a vanilla SGD optimiser for training these models with a learning rate $\eta$ (a hyperparameter selected from $\{10^{-3}, 10^{-4}\}$ as described in further detail below) and a momentum of $0.9$.
We aim to maximise our hardware utilisation by using a batch size that is as large as possible to fit the model's weights, activations, and gradients during training in the memory of a single of our GPUs.
Moreover, to simplify our implementation of TAB within the Pytorch ecosystem, we look for batch sizes that are divisors of the training set (as this enables us to very easily store a model's loss history during training).
We note, however, that this is not in any way a hard requirement for TAB but rather something we exploited to easily adapt it to a commonly used framework such as Pytorch.
This process resulted in using the following batch sizes for our real-world datasets: $B = 137$ for \texttt{Waterbirds}, $B = 436$ for \texttt{CelebA}, $B = 194$ for \texttt{BAR}, and $B = 128$ for \texttt{CUB}.

For all models and tasks, we train models for a total of maximum $T$ epochs with $T = 100$ for the synthetic tasks, $T = 300$ for \texttt{Waterbirds}, $T = 150$ for \texttt{CelebA}, $T = 200$ for \texttt{BAR}, and $T = 300$ for \texttt{CUB}.
We stop training before this limit using early stopping based on the validation accuracy.
In this setup, we interrupt training if a model's validation accuracy does not increase by more than $0.001$ (``\textit{stopping delta}'') in the last $5$ epochs (``\textit{patience''}).
Finally, to help find better optima across all tasks and models, we reduce the learning rate by $\times 0.1$ when the training loss plateaus for $10$ epochs.

\subsection{Model Selection}
\label{sec:appendix_model_hyperparameters}

To conduct a fair evaluation of all baselines, we perform an extensive hyperparameter search for all methods.
Here, we discuss which hyperparameters we search over for all methods, as well as the hyperparameters selected for each task after averaging the results over three distinct runs.
This hyperparameter selection is done by performing a grid search over all combinations of hyperparameter candidates and selecting the model with the \textit{highest validation accuracy}.
The only exception for this is G-DRO, for which we perform model selection based on validation WGA given that we assume group labels are provided during training and we use G-DRO as an upper bound for unsupervised BM.
Finally, to further reduce the computational cost of our model selection, the learning rate $\eta$ and weight decay $\lambda_{\ell_2}$ hyperparameters used for all non-ERM and non-G-DRO methods are set to those selected for their ERM equivalent (i.e., we first train the ERM baseline, find the best learning rate $\eta$ and weight decay $\lambda_{\ell_2}$ and fix those when performing model selection and training of all other baselines).
Below we describe the hyperparameters we searched over for each baseline.

\indent\newheader{ERM - $\{\eta, \lambda_{\ell_2}\}$}
When training ERM models, we select their learning rates from $\eta \in \{10^{-3}, 10^{-4}\}$ and their weight decay factors from $\lambda_{\ell_2} \in \{$0, 0.0001, 0.01, 1$\}$  (with $0.00001$ added for the MNIST-based datasets as they are smaller and allow for larger searches).
This results in a total of $2 \times 4 \times 3 = 24$ models trained per task to get the mean validation accuracy of each setup across three different random seeds.
When selecting models based on validation accuracy, we got the following configuration for all tasks: $(\eta, \lambda_{\ell_2}) = (10^{-3}, 0)$ for \texttt{Even-Odd}, $(\eta, \lambda_{\ell_2}) = (10^{-3}, 0.0001)$ for \texttt{cMNIST}, $(\eta, \lambda_{\ell_2}) = (10^{-3}, 0)$ for \texttt{BAR},  $(\eta, \lambda_{\ell_2}) = (10^{-3}, 0.01)$ for \texttt{Waterbirds} and $(\eta, \lambda_{\ell_2}) = (10^{-3}, 0.0001)$ for \texttt{CelebA}.
For \texttt{CUB}, we observed that $\eta \in \{10^{-3}, 10^{-4}\}$ performed poorly, so we added $\eta = 10^{-2}$ as a candidate to this set.
This resulted in us selecting  $(\eta, \lambda_{\ell_2}) = (10^{-2}, 0.0001)$ for \texttt{CUB}.

\indent\newheader{G-DRO~\cite{hu_gdro_og, sagawa_gdro_bias_mitigation_19} - $\{\eta, \lambda_{\ell_2}\}$}
As in ERM, for G-DRO we carefully select the learning rate and the weight decay as both of these hyperparameters have been shown to be highly important for good worst-group performance~\cite{sagawa_gdro_bias_mitigation_19}.
Therefore, we select hyperparameters by searching over candidate learning rates $\eta \in \{10^{-3}, 10^{-4}\}$ and candidate weight decays $\lambda_{\ell_2} \in \{0, 0.0001, 0.01, 1\}$.
Moreover, we use the stable online G-DRO algorithm by Sawaga et al.~\cite{sagawa_gdro_bias_mitigation_19}, with an exponential decay factor of $\gamma = 0.01$, as this approach has been shown to be more stable in practice, particularly when minority groups are small.
Our hyperparameter grid search yields the following hyperparameters across the different group-annotated tasks: $(\eta, \lambda_{\ell_2}) = (10^{-3}, 0.0001)$ for \texttt{Even-Odd}, $(\eta, \lambda_{\ell_2}) = (10^{-3}, 0)$ for \texttt{cMNIST}, $(\eta, \lambda_{\ell_2}) = (10^{-3}, 0.1)$ for \texttt{Waterbirds}, and $(\eta, \lambda_{\ell_2}) = (10^{-3}, 0.01)$ for \texttt{CelebA}.

\indent\newheader{LfF~\cite{nam_lff_bias_mitigation_neurips_20} - $\{q\}$}
LfF uses a generalised cross-entropy loss~\cite{zhang2018generalized} to dynamically learn a biased model for weighting the loss of an ``unbiased'' model.
This loss depends on a hyperparameter $q \in (0, 1]$ that controls the level of bias amplification of the biased model (the loss becomes a vanilla cross-entropy loss as $q \rightarrow 0$).
Therefore, across all tasks, we attempted values $q \in \{$0.05, 0.1, 0.25, 0.5, 0.75, 0.9, 0.95$\}$ for LfF.
This yields the following hyperparameters across our different tasks: $q = 0.05$ for \texttt{Even-Odd}, \texttt{cMNIST}, \texttt{Waterbirds}, \texttt{CelebA}, and \texttt{CUB}, and $q = 0.1$ for \texttt{BAR}.
We note that, as opposed to the hyperparameters selected when using best validation accuracy, the optimal configurations for LfF (i.e., the hyperparameters with the highest test loss) are more diverse: $q = 0.1$ for \texttt{Even-Odd} and \text{CUB}, $q = 0.75$ for \texttt{cMNIST} and \texttt{Waterbirds}, $q = 0.5$ for \texttt{CelebA}, and $q = 0.95$ for \texttt{BAR}.
These results, therefore, show how determining the right value for $q$ can be very hard in practice.

\indent\newheader{JTT~\cite{liu_jtt_bias_mitigation_icml_21} - $\{T, \lambda_{\text{up}}\}$}
As discussed in Section~2, JTT has two hyperparameters: $T$, the number of epochs one trains the identifier model for, and $\lambda_{\text{up}}$, how much we should upweight each mispredicted sample by the identifier model.
Following the same official implementation by Liu et al.~\cite{liu_jtt_bias_mitigation_icml_21}, we perform the upweighting in JTT by constructing a new dataset in which each mispredicted training sample is included $\lambda_{\text{up}}$ times.
This enables significantly better results and much more stable training.
However, it comes with the added cost of significant training times as this new dataset could be as large as $\lambda_{\text{up}} N$ in the worst case.
Following similar values used by the original authors of this work, we perform a grid search over $T \in \{1, 5, 10, 25, 50\}$ and $\lambda_{\text{up}} \in \{10, 25, 50, 100, \text{``ratio''} \}$, where ``ratio'' indicates a dynamic computation of $\lambda_{\text{up}}$ whereas we set this value to the ratio between the size of the set of samples correctly predicted by the identifier model and the size of the set of samples mispredicted by the identifier model.
The only exception for this is in CUB, where we only search over $T \in \{$10, 25, 50 $\}$ and $\lambda_{\text{up}} \in \{$ 10, 25 $\}$ as otherwise training times got intractable.
This process yielded the following selected hyperparameters across all tasks: $(T, \lambda_{\text{up}}) = (50, 10)$ for \texttt{Even-Odd}, $(T, \lambda_{\text{up}}) = (50, 25)$ for \texttt{cMNIST}, $(T, \lambda_{\text{up}}) = (10, 25)$ for \texttt{Waterbirds}, $(T, \lambda_{\text{up}}) = (10, 1)$ for \texttt{CelebA}, $(T, \lambda_{\text{up}}) = (50, 25)$ for \texttt{BAR}, and $(T, \lambda_{\text{up}}) = (25, 25)$ for \texttt{CUB}.

\indent\newheader{MaskTune~\cite{asgari_masktune_bias_mitigation_neurips_22} - $\{\tau\}$}
MaskTune operates by (1) first training a model via ERM, (2) then generating a new dataset by masking each sample $\mathbf{x}^{(i)}$ based on the saliency map~\cite{og_saliency} of that sample by the ERM model, and (3) finally fine-tuning the ERM model for a single epoch using a small learning rate on the newly constructed dataset.
The intuition of this approach is that by masking areas of the image that a model is attending to, one may get rid of easily exploitable spurious correlations and force the model to learn to make a prediction using alternative features (i.e., by learning to generalise).
Assuming that (1) we always fine-tune for a single epoch (as recommended by the authors), (2) we use X-GradCam~\cite{fu2020axiom} as the underlying saliency method (as used in the original paper), and (3) we set the fine-tuning learning rate to be the end learning rate of the original ERM model 
(as suggested by the authors), the only hyperparameter in MaskTune we control is $\tau \in \mathbb{R}$, the threshold controlling which pixels to mask in the saliency map of each sample.
Following the values for $\tau$ used in MaskTune's original paper, we try values of $\tau$ in $\tau \in \{\mu, \mu + 0.5\sigma^2, \mu + \sigma^2, \cdots, \mu + 2.5\sigma^2, \mu + 3\sigma^2\}$ where $\mu$ is the mean of the saliency map pixels of a specific sample and $\sigma^2$ is the standard deviation (i.e., the actual threshold is a function of the sample's saliency map).
This process results in us selecting the following hyperparameters across our tasks: 
$\tau = \mu + 3\sigma^2$ for \texttt{Even-Odd}, $\tau = \mu + 2.5\sigma^2$ for \texttt{cMNIST}, $\tau = \mu$ for \texttt{Waterbirds}, $\tau = \mu + 2.5\sigma^2$ for \texttt{CelebA}, $\tau = \mu + 1.5\sigma^2$ for \texttt{BAR}, and $\tau = \mu + 3\sigma^2$ for \texttt{CUB}.

\indent\newheader{TAB - $\emptyset$}
As discussed in Section~5, our method TAB has no hyperparameters and, therefore, no need for fine-tuning.
Therefore, we deployed TAB without changing any of the hyperparameters used for the equivalent ERM model with the exception that when training TAB's identifier model to collect training loss histories, we perform early \textit{stopping based on the training accuracy} rather than the validation accuracy.
We do this so that training of the identifier stops once the model has perfectly fitted the training set, at which point additional losses in the history come with diminishing returns.

\subsection{Software and Hardware Used}
\label{appendix:software_and_hardware_used}

\newheader{Software}
For this work, we constructed our codebase based on the MIT-licensed open-source repository by Espinosa Zarlenga et al.~\cite{cem, intcem}, which provided a useful starting point for easy experiment tracking and deployment.
Our implementation of TAB and all underlying DNNs is built on top of PyTorch 1.12~\cite{pytorch}, a commonly used deep learning library with a BSD license.
For G-DRO, we adapted the official implementation of this method by Sawaga et al.~\cite{sagawa_gdro_bias_mitigation_19} by updating the code so that it could run within our infrastructure while maintaining all key pieces as in their MIT-licensed \href{https://github.com/kohpangwei/group_DRO}{public repository}.
For LfF, we adapted the authors' \href{https://github.com/alinlab/LfF}{official implementation} into our own infrastructure, keeping the setup and generalised cross-entropy computation intact.
For JTT, we reimplemented their training algorithm based on their \href{https://github.com/anniesch/jtt}{official implementation}.
Finally, for MaskTune, we ported the authors' \href{https://github.com/aliasgharkhani/Masktune}{official implementation} into our own infrastructure.

All figures with the exception of Figure~1 were generated using Matplotlib 3.5, a BSD-licensed Python plotting library.
Figure~1 was instead generated using \href{https://github.com/jgraph/drawio}{draw.io}, an open-sourced drawing software distributed under an Apache 2.0 license.
All of the code, configs, and scripts needed to recreate our results, will be made public through an open-source repository upon publication of this paper.

\newheader{Resources}  We ran all of our experiments on two compute clusters.
The first cluster consisted of a machine with four Nvidia Titan Xp GPUs and 40 Intel(R) Xeon(R) E5-2630 v4 CPUs.
The second cluster consisted of a machine with a single Nvidia Quadro RTX 8000 GPU and eight Intel(R) Xeon(R) Gold 5218 CPUs.
We estimate that all experiments, including initial explorations and all the model selections we ran, required between 450 and 500 GPU hours.

\section{Details for Motivation Experiments}
\label{appendix:motivation_experiment_details}

In this section we provide details on the constructions of the figures used in Section~4, where we describe the importance of validation group annotations for model selection in BM pipelines.
Figures~2, 3, and 4 were all generated by training a \textit{single} Resenet-34 model for $50$ epochs.
This model is trained on our \texttt{Waterbirds} task for all figures and on the \texttt{CelebA} task for Figure~4.
We follow the same approach as described in Appendix~\ref{appendix:hyperparameters} to train this model in each respective task with the difference that, for the models shown in Figures~2 and 4, we do not perform early stopping (as we want to emphasise how biases in the validation set can in fact lead to an unwanted early stopping).
Notice that in Figure~2 we use colours to indicate points in a curve that are higher (green) or lower (red) than the the metric at the beginning of the marked grey zone.
Finally, in Figure~4 we split samples between bias-conflicting and bias-aligned by identifying minority groups lacking the spurious correlation as bias-conflicting.
In \texttt{Waterbirds} we use (``waterbirds on land backgrounds'' and ``landbirds on water backgrounds'') as our bias-conflicting groups, as they are both similarly underrepresented and lacking the spurious correlation, while in \texttt{CelebA} we use ``not blonde male faces'' as the bias-conflicting slice shown (as they are severely underrepresented and lacking the exploitable gender-based spurious correlation).

\section{Random Transformations to Augmented Samples}
\label{appendix:transformation}

In this section, we explore how domain-specific knowledge can be used to improve TAB's performance even further through the use of augmentation transformations.
For this, when we construct TAB's augmented set $\mathcal{D}_l^\prime$, we apply an stochastic transformation function $\eta: \mathbb{R}^n \rightarrow \mathbb{R}^n$ to each sample (i.e., $\mathcal{D}_l^\prime= \{(\eta(\mathbf{x}^{\prime(i)}), y^{\prime(i)}) \; | \; (\mathbf{x}^{\prime(i)}, y^{\prime(i)}) \in \mathcal{D}_l^-\}_{i=1}^z$).
We explore this in our MNIST-based tasks as we know handwritten digits should be invariant to small perturbations in their angles and locations.
Specifically, we let $\eta$ be a transformation pipeline that applies small random rotations ($\theta \sim \text{Unif}([-\pi/6, \pi/6])$), translations ($\Delta x, \Delta y\sim \text{Unif}([0, l/10])$), and Gaussian blurring (using a 5x5 filter).

\begin{figure}[h!]
    \centering
    \includegraphics[width=\textwidth]{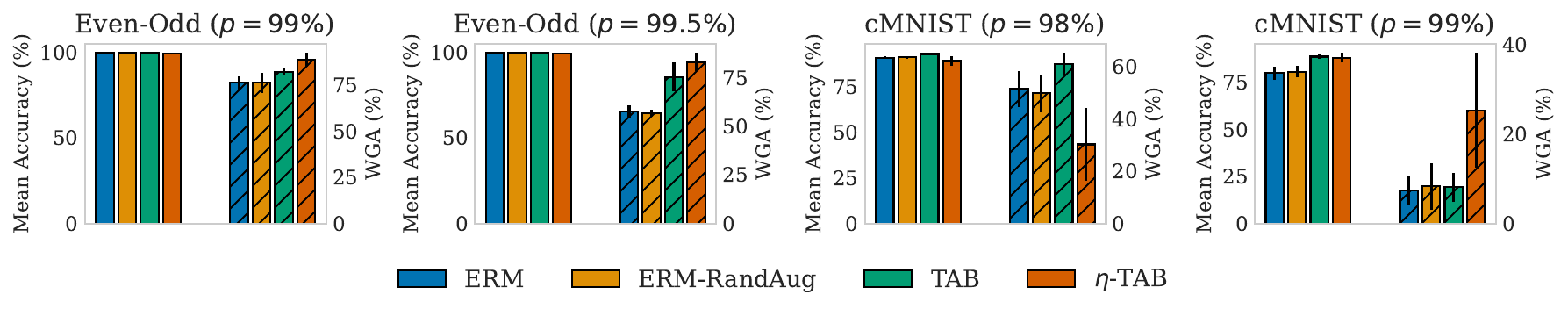}
    \caption{
        Mean accuracy (left group of solid bars) and WGA (right group of dashed bars) of TAB with and without random transformations of upsampled examples in \texttt{Even-Odd} and \texttt{cMNIST} as we vary $p$ for both datasets.
        As baselines, we show the performance of an equivalent ERM model trained both with and without the same random transformations for its training set.
    }
    \label{fig:transformations}
\end{figure}

Our results, shown in Fig.~\ref{fig:transformations}, show that in highly imbalanced domains (e.g., $p = 0.99$ in \texttt{cMNIST} and $p \in \{0.99, 0.995\}$ in \texttt{Even-Odd}), TAB's WGA can be significantly improved by adding domain-specific transformations to their augmented samples without a clear sacrifice in mean accuracy.
Specifically, TAB's version including the transformations ($\eta$-\textit{TAB}) achieves up to a $17\%$ absolute improvement in WGA over the standard TAB pipeline for \texttt{cMNIST} when $p = 0.99$.
Moreover, we see that these same benefits cannot be harvested by simply randomly applying the same transformations to the original training set during training.
This is shown by the similar performance between the original ERM model and \textit{ERM-RandAug}, an ERM model trained on a dataset where samples are randomly augmented using the same transformations as for $\eta$-TAB during training.

In contrast to the boost obtained by applying target transformations, we notice that such transformation may result in worse WGA when the worst-group imbalance is not extreme (e.g., $p = 0.98$ for \texttt{cMNIST}).
Hence, although some of our preliminary results here suggest that these transformations can be useful in extremely imbalanced setups, more work is needed to understand (1) how to correctly design these transformations when domain knowledge is available, (2) how to determine whether such transformations can be properly applied without damaging a model's worst-group performance, and (3) what are some of the repercussions on TAB's PoU of introducing new hyperparameters coming from such transformation pipelines.
We believe these to be promising and important directions for future work.

\section{Effect of Batching Clustering for TAB}
\label{appendix:batched_knn}
\begin{figure}[h!]
    \centering
    \includegraphics[width=\textwidth]{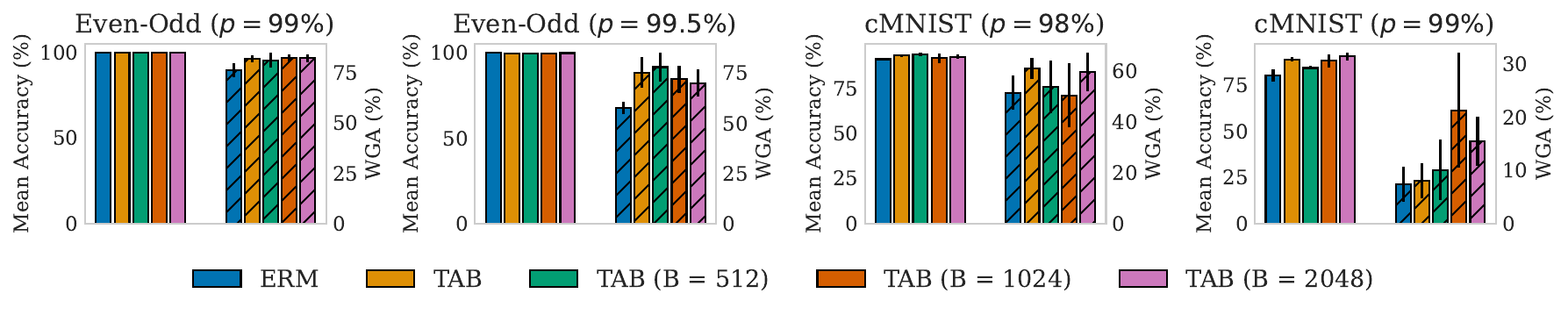}
    \caption{
        Mean accuracy (left group of solid bars) and WGA (right group of dashed bars) of TAB when using mini-batched $k$-means clustering for its dataset partition step.
        We show our results in \texttt{Even-Odd} and \texttt{cMNIST} while we vary $p$ to simulate extremely biased setups.
        As baselines, we show the performance of equivalent ERM and TAB models.
    }
    \label{fig:minibatching}
\end{figure}

In this section, we explore the effect of using mini-batched $k$-means~\cite{batch_kmeans} for clustering history losses in TAB's pipeline.
Exploring how mini-batching can be used as a part of TAB can be insightful to understanding how our method may be able to scale to datasets where $k$-means may become a bottleneck (i.e., datasets with a size in the order of millions or billions of samples).
With this goal in mind, in Figure~\ref{fig:minibatching} we show the effect of using batches for $k$-means in our synthetic datasets. For these results, we use Scikit-learn's official MiniBatched-$k$-Means~\cite{scikit-learn} implementation with its default settings except for the ``reassignment ratio'', which we decrease to $0.00001$ to enable small clusters to be discovered (as we operate under the assumption that one of the clusters will be significantly smaller).

Our results show that mini-batching $k$-means enables TAB to still learn less biased models than standard ERM models.
This suggests that our method can still be easily deployed in practice to learn models that are less biased than their ERM equivalents in extremely large datasets without the need for any model selection.
However, we also notice that batching comes with a significant variance that can, in some instances, lead to bad clusters being discovered.
This results in a loss on some of the benefits we were able to obtain from TAB's original full dataset clustering.
We hypothesise that this is due to the fact that, in these extremely group-imbalanced datasets, it is likely that no spurious-conflicting samples may be found in a given batch when learning clusters using mini-batched $k$-means.
Therefore,
future work may focus on understanding how to fully recover TAB's worst-group performance when using only batches or subsets of the training set when discovering bias-conflicting and bias-aligned subgroups.

\end{document}